\documentclass[aoas]{imsart}

\RequirePackage{amsthm,amsmath,amsfonts,amssymb}
\RequirePackage[authoryear]{natbib}
\usepackage{amsmath,amsthm,amssymb,bm,mathtools}
\usepackage{graphicx,psfrag,epsf,algorithm,algpseudocode}
\usepackage{enumerate,multirow,array,color,natbib,url,booktabs}
\DeclarePairedDelimiter\floor{\lfloor}{\rfloor}
\usepackage[font=small,labelfont=bf]{caption}
\usepackage{adjustbox}
\usepackage{subfigure}

\newtheorem{thm}{Theorem}

\newtheorem{lemma}{Lemma}

\let\proglang=\textsf

\newcommand{\Mean}{{\mathbb{E}}}
\newcommand{\Var}{{\mbox{Var}}}

\newcommand{\prob}{{\mbox{Pr}}}

\DeclareMathOperator*{\argmin}{arg\,min}

\def\spacingset#1{\renewcommand{\baselinestretch}%
	{#1}\small\normalsize} \spacingset{1}

\usepackage{xr}
\startlocaldefs

\endlocaldefs

\begin{document}
	
	\begin{frontmatter}
		\title{A Multi-Agent Reinforcement Learning Framework for Off-Policy Evaluation in Two-sided Markets}
		\runtitle{MARL for Off-Policy Evaluation in Two-sided Markets}
		
		
		\begin{aug}
			\author[A]{\fnms{Chengchun} \snm{Shi}\ead[label=e1]{c.shi7@lse.ac.uk}}
			\author[B]{\fnms{Runzhe} \snm{Wan}\ead[label=e2]{rwan@ncsu.edu}}
			\author[C]{\fnms{Ge} \snm{Song}\ead[label=e3]{songge@didiglobal.com}}
			\author[C]{\fnms{Shikai} \snm{Luo}\ead[label=e4]{luoshikai@didiglobal.com}}
			\author[D]{\fnms{Hongtu} \snm{Zhu}\ead[label=e6]{htzhu@email.unc.edu}}
			\author[B]{\fnms{Rui} \snm{Song}\ead[label=e5]{rsong@ncsu.edu}}
			\address[A]{London School of Economics and Political Science, \printead{e1}}
			\address[B]{North Carolina State University, \printead{e2,e5}}
			\address[C]{Didi Chuxing, 
				\printead{e3,e4}}
			\address[D]{The Univeristy of North Carolina at Chapell Hill,
				\printead{e6}}
		\end{aug}
		\begin{abstract}
			The two-sided markets such as ride-sharing companies often involve  a group of subjects who are making sequential decisions across	time and/or location.  With the rapid development of smart phones and internet of things, they have substantially transformed the transportation landscape of human beings.  In this paper we consider  large-scale fleet management in ride-sharing companies that involve  multiple units in different areas receiving sequences of products (or treatments) over time.  Major technical challenges, such as {\color{black}policy evaluation},   arise in those  studies because (i) spatial and temporal proximities induce interference between locations and times; and (ii) the large number of locations results in the curse of dimensionality. {\color{black}To address both challenges simultaneously,} we introduce a multi-agent reinforcement learning (MARL) framework for carrying {\color{black}policy evaluation} in these studies. We propose novel estimators for mean outcomes under different products that are consistent despite the high-dimensionality of state-action space. The proposed estimator works favorably in simulation experiments. We further illustrate our method using a real dataset obtained from a two-sided marketplace company  to evaluate the effects of applying different subsidizing policies. A {\proglang{Python}} implementation of our proposed method is available at \url{https://github.com/RunzheStat/CausalMARL}. 
		\end{abstract}
	\end{frontmatter}

\section{Introduction}\label{secintroduction}


This paper concerns the applications in the two-sided markets that involve  a group of subjects who are making sequential decisions across
time and/or location.  
In particular, we consider large-scale fleet management in ride-sharing companies, such as Uber, Lyft and Didi. These companies form a typical two-sided market that enables efficient interactions between passengers and drivers \citep{armstrong2006competition,Rysman2009}. With the rapid development of smart phones and internet of things, they have substantially transformed the transportation landscape of human beings \citep{Frenken2017,Jin2018,Hagiu2019}. With rich information on passenger demand and locations of taxi drivers, they significantly reduce taxi cruise time and passenger waiting time in comparison to traditional taxi systems \citep{li2011hunting,zhang2014understanding,miao2016taxi}. {\color{black}We use the numbers of drivers and call orders to measure the supply and demand at a given time and location. Both supply and demand are spatio-temporal processes and they interact with each other. These processes depend strongly on the platform's policies, and have a huge impact on the platform's outcomes of interest, such as drivers' income level and working time, passengers' satisfaction rate, order answering rate and order finishing rate, etc.}

A fundamental question of interest that we consider here is how to establish causal relationships between platform policies and platform's outcomes. In particular, we are interested in evaluating the causal effects of applying different subsidizing policies or recommendation programs to drivers or passengers in different spatial locations of a city. The purpose of implementing these policies is to balance the taxi supply and passenger demand across different areas of the city, so as to meet more passengers' requests and reduce drivers' vacant time. As an example, suppose a passenger opens the ride-sharing application on their smart phone and enters their destination. 
The platform will decide whether to 
recommend the customer to join a program and
send them a coupon to discount this ride, 
depending on their locations. 
Such a recommendation increases the chance that the customer orders this particular ride to reduce the local drivers' vacant time. As another example, suppose the ride-sharing platform provides subsides to drivers in areas where there are more passenger call orders than the number of drivers. Implementing such a policy will attract more drivers to these areas, meeting more passengers' requests. 

Solving this fundamental question faces at least two major challenges. The first one is that the spatial and temporal proximities in the aforementioned applications will induce interference between locations and times. {\color{black}As such, the outcome associated with each unit might depend on the treatments of all regions. Learning each spatial unit's value based on its own data only would yield a biased estimator. See e.g., the performance of the baseline estimator DR-NS in Sections \ref{sec:numerical} and \ref{sec:real}. We notice that most of the existing policy evaluation methods in the literature focus on the setting where no interference occurs, that is, the outcome of each experimental unit is unaffected by treatment assignment of other units \citep[see e.g.,][]{zhang2012robust,chakraborty2014inference,dudik2014doubly,matsouaka2014evaluating,luedtke2016statistical,belloni2017program,wu2020resampling,shi2020breaking}.} Such a no interference assumption is often referred to as the stable unit treatment value assumption  \citep[SUTVA,][]{rubin1980randomization,rubin1986comment} in the causal inference literature. To elaborate the violation of SUTVA, let us revisit the example of applying driver-side subsidizing policies. As we have commented, applying a subsidizing policy at one location would attract some drivers from its  neighbouring   areas to that location,  so  the subsidizing policy at one location could influence various  outcomes of those neighbouring  areas, inducing interference among spatial units. Moreover, the subsidizing policy at a given time would  affect both current and future supplies, inducing interference across time. In the passenger recommendation program example, sending coupons to certain passengers not only increases the number of current call orders, but the chance that these passenger use the app more frequently in the future as well. As such, applying the recommendation program will increase the number of future call orders, inducing interference across time. 
Thus, interference across time and/or among spatial units leads to the violation of SUTVA in our applications. 


\begin{figure}[!t]		
	\centering
	\includegraphics[width=0.4\linewidth]{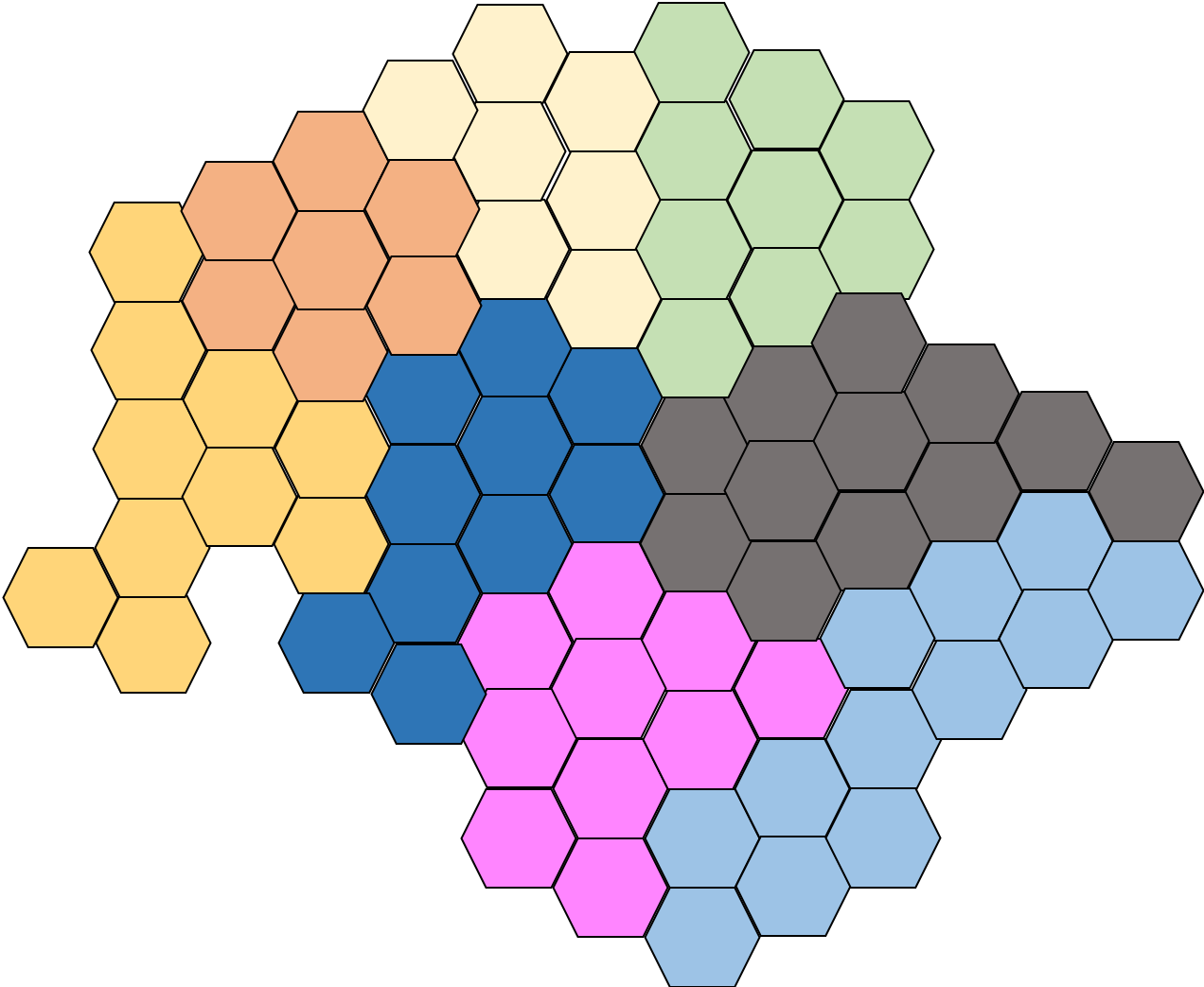}
	\caption{Visualization of $N=8$ different spatial units in the city.}\label{fig0}
\end{figure}

The second challenge is that the large number of spatial units results in the curse of dimensionality. Specifically, to implement the subsidizing policy or the customer recommendation program, the whole city is divided into $N=8$ disjoint spatial units, as shown in Figure \ref{fig0}. For each spatial unit, the company can decide whether to apply certain promotion strategy to this region or not. {\color{black}In the presence of spatial interference, the amount of data needed to provide reliable policy value estimates grows exponentially with $N$, resulting in the curse of dimensionality. It remains challenging to effectively model the high-dimensional system without additional assumptions.}

The aim of this paper is to evaluate the treatment effects of multiple policies in the presence of spatiotemporal interference. There is a huge literature on causal inference, but most existing works assume SUTVA \citep[see e.g.,][]{hirano2003efficient,Imbens2015, Wager2018,Yao2020}. There has been substantial interest in the development of causal inference under interference. Our work falls into an emerging  research topic  on space- or time-dependent treatment effects evaluation \citep[see e.g.,][]{Hudgens2008,Eric2012,toulis2013estimation,halloran2016dependent,dempsey2017stratified,Athey2018,Murphy2018,bhattacharya2019causal,Bojinov2019,ning2019Bayes,Reich2020}. However, none of the above cited works studies the interference effects in both space and time. In particular, \cite{Reich2020} gave a systematic review of various statistical models for spatial causal inference and pinpoint some areas of future work. However, those models in \cite{Reich2020} were primarily motivated by research questions in environmental and epidemiological studies, so their generalization to two-sided markets remains unknown.

Reinforcement learning is a general machine learning technique that allows an agent to interact with a given environment,  which has drawn more and more attention in the statistics literature, see \citet{Sutton2018} for an overview. 
Recently, a number of proposals utilize reinforcement learning in mobile health or two-sided markets  \citep{ertefaie2014,luckett2019,chen2020heterogeneous,hu2020personalized,liao2019off,wang2021projected,zhou2021estimating,li2022reinforcement,li2022rate,liao2020batch,shi2022dynamic,shi2021statistical}. In addition, there is a growing literature on adapting reinforcement learning to develop dynamic treatment regimes in precision medicine, to recommend treatment decisions based on individual patients' information \citep{Murphy2003,chak2010,qian2011performance,zhao2012,zhang2013,song2015penalized,zhao2015,zhangyc2015,zhangyc2016,zhu2017greedy,wang2018quantile,shi2018high,shi2018maximin,mo2020learning,meng2020near,cai2021deep,fang2022fairness}.
All these methods considered a single-agent setup where only one agent exists in the environment. 

The proposed method is motivated by a line of research on multi-agent reinforcement learning (MARL) in the cooperative setting \citep[see e.g.,][for an overview]{zhang2019multi}. 
In MARL, multiple autonomous agents operate in a common environment. Each agent aims to maximize its own cumulative reward by interacting with the environment and other agents. Compared to the single-agent setup, MARL is much more challenging due to the presence of the high-dimensional action space induced by  multiple agents. 
Most existing works on MARL focus on the \textit{policy optimization} problem where each agent aims to identify an optimal policy that optimizes its long-term reward. In particular, \cite{yang2018mean} developed a mean field Q-learning algorithm in the discounted-reward setting. 
We remark that many of these methods are not directly applicable to the problem of {\textit{policy evaluation}}, where the objective is to learn the impact of a given policy using data collected possibly from a different behavior policy. 

\begin{figure}[!t]
	\centering
	\includegraphics[width=12cm]{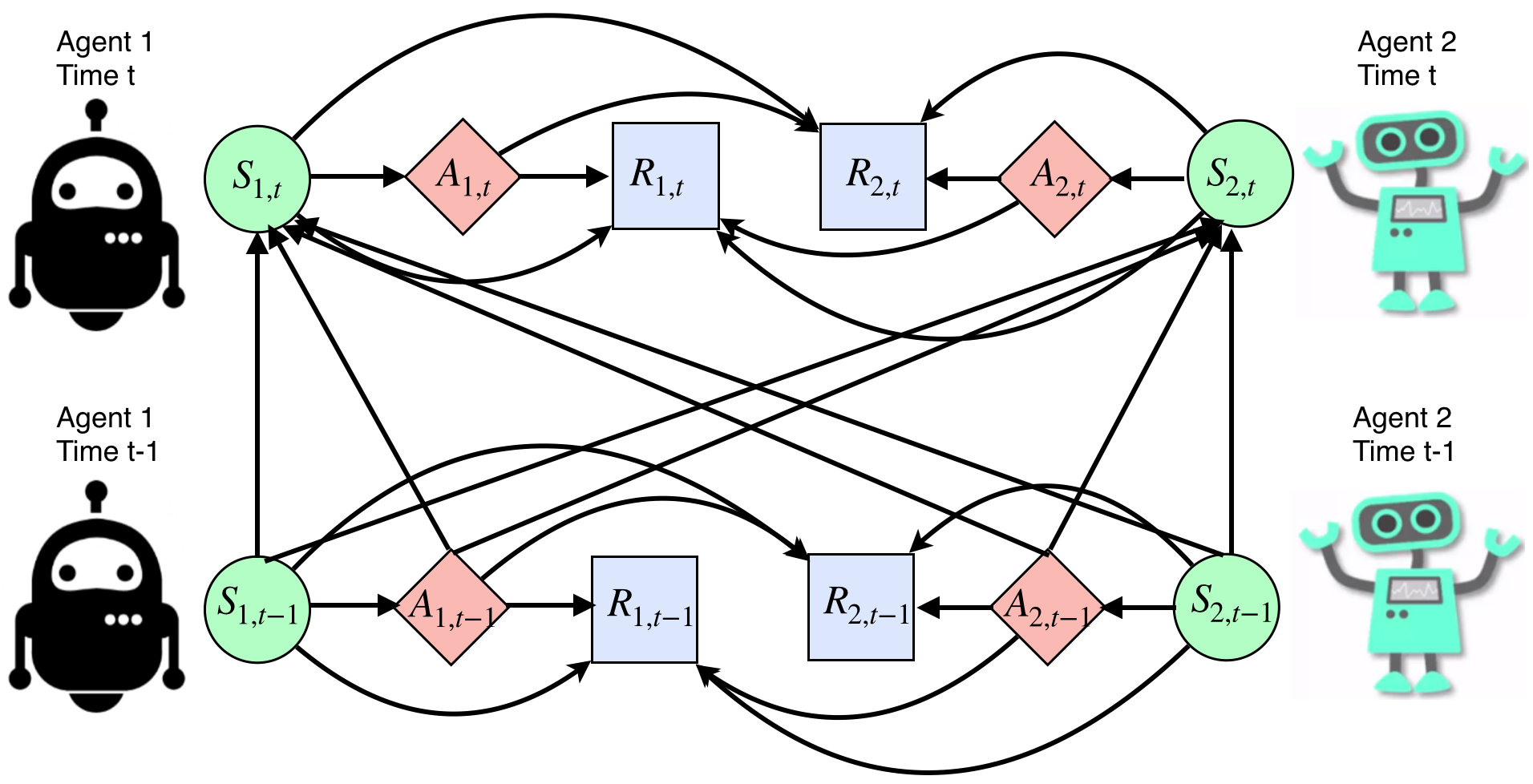}
	
	\caption{Causal diagram for a multi-agent system with two agents. $(S_{j,t},A_{j,t},R_{j,t})$ represents the state-treatment-outcome triplet of the $j$-th agent at time $t$.}\label{fig1}
\end{figure}

The contributions of this paper are summarized as follows. First, we introduce a multi-agent reinforcement learning framework for {\color{black}policy evaluation}. Consider the example of applying driver-side subsidizing policies or passenger recommendation programs. Each spatial unit in the city is considered as an agent. In addition to the treatment-outcome pairs, it is assumed that each agent is associated with a set of time-varying confounding variables. This naturally leads to a multi-agent system. Under this framework, the interference effects in space are modeled by the interactions between different agents, while  the interference effects in time are modeled by the dynamic system transitions. See the causal diagram depicted in Figure \ref{fig1} for an illustration. 
Estimation of the mean outcome under different products is reduced to the off-policy evaluation problem in MARL. To the best of our knowledge, this is the first work that explores MARL in the statistics literature. {\color{black}We remark that alternative to the MARL framework, the policy evaluation problem can be equivalently formulated using a single-agent setup whose action space is $\{0,1\}^N$ and reward is the sum of individual rewards of the agents. However, as mentioned earlier, it remains challenging to handle the exponentially large action space under the single-agent formulation. To the contrary, our formulation allows us to borrow ideas from the MARL literature to simultaneously address the challenges of dealing with spatiotemporal interference as well as the curse of dimensionality.}

Second, we propose an original off-policy policy evaluation procedure in MARL. 
A number of off-policy evaluation algorithms have been developed under the single-agent setup \citep[see e.g.,][]{thomas2015,jiang2016,thomas2016data,liu2018,kallus2019efficiently,tang2019doubly,uehara2019minimax,shi2021deeply,chen2022well}. {\color{black}Directly applying these methods will yield value estimators with large variances, due to the curse of dimensionality. See the performance of the baseline estimator and DR-NM in Sections \ref{sec:numerical} and \ref{sec:real}. 
	Our proposal is the first to harness the power of modern MARL and single-agent policy evaluation methods, while establishing rigorous statistical guarantees, when tackling an important business question in two-sided markets. The proposed estimator requires estimation of the density ratio of the stationary state distribution and the Q-function associated with each single agent. The key ingredient of our method lies in learning both the density ratio and Q-function based on the mean-field approximation and aggregating these estimators properly to satisfy the doubly-robustness property in the average reward setting. Event though ideas such as the mean-field approximation and doubly robust estimation are not completely new, how to integrate them properly and effectively into a value estimator with desired theoretical guarantees is nontrivial, and is one of the main contributions of this article. In particular, we propose an original mean-field algorithm to approximate the density ratio in MARL. We also extend \cite{yang2018mean}'s proposal to the average-reward setting, which is more suitable for our purpose of policy evaluation. 
	The mean field approximation effectively reduces the high-dimensional state-action space to a moderate scale, leading to a value estimator with decreased variance. The doubly-robustness guarantees that  our estimated value is consistent when either the density ratio or the Q-function is well-approximated, reducing its bias resulting from the mean-field approximation. }

{\color{black}Third, we systematically study how to test the mean-field approximation assumption from the observed data. Specifically, we show that under an additive noise model assumption, the mean-field approximation assumption is reduced to certain conditional mean independence assumptions. See Appendix \ref{app:moremf} for details. This observation allows us to employ existing state-of-the-art conditional independence tests to examine the validity of mean-field approximation in practice. To our knowledge, this is the first time that a formal statistical test for such an assumption is proposed. We illustrate the idea using the forward-backward learning-based test developed by \cite{shi2020does} in our numerical studies. }

Finally, we investigate the statistical properties of our estimator. 
In particular, we establish its doubly-robustness property (Theorem \ref{thm:double}) and derive its ``oracle" property when both the density ratio and the Q-function are well approximated (Theorem \ref{thm:oracleest}). {\color{black}Our theory allows the number of spatial units, denoted as $N$,  to be either bounded or diverge to infinity. Therefore, the proposed estimator offers a useful policy evaluation tool to a wide range of applications in the presence of spatiotemporal interference. To prove these results, we develop an exponential inequality for the suprema of empirical processes under weak dependence (Lemma \ref{lemma:EP}), which is useful for finite-sample analysis of machine learning estimates based on dependent observations. As we have mentioned earlier, most off-policy evaluation algorithms are developed under a single-agent setup. To our knowledge, these theoretical results have not been established in the existing literature.}

The rest of the article is organized as follows. In Section 
\ref{sec:causalMARL}, we describe the problem setup and introduce a potential outcome framework for MARL. We present our method in Section \ref{sec:method}. Simulation studies are conducted in Section \ref{sec:numerical}. In Section \ref{sec:real}, we apply the proposed method to a dataset from a two-sided marketplace company to evaluate the effects of applying different subsidizing policies. Finally, we conclude our paper by a discussions section. Statistical property of our method is investigated in the appendix. 




\section{A potential outcome framework for MARL}\label{sec:causalMARL}
In this section, we extend Rubin's potential outcome framework to the multi-agent system. This allows us to formulate our causal estimand. We first introduce some notations.  For $1\le i\le N$, we  consider  two treatments (actions) associated with the $i$-th spatial unit (agent)  such that  the action space is represented as  $\mathcal{A}_i=\{0,1\}$. Let $\mathbb{S}_i$ denote the state space associated with the $i$-th agent. In our application, the two treatments correspond to applying certain promotion strategy to a given spatial unit or not. The state variables include the number of call orders (demand) and available drivers (supply), and a  supply-demand equilibrium metric that measures the degree of mismatch between orders and drivers within each region.
The joint state and action spaces are given by $\mathbb{S}= \mathbb{S}_1 \times \mathbb{S}_2 \times \ldots \times \mathbb{S}_N$ and $\mathcal{A}=\mathcal{A}_1\times \mathcal{A}_2 \times  \ldots \times \mathcal{A}_N=\{0,1\}^N$, respectively.  
For a sequence of $N$-dimensional vectors $\bm{a}_0,\bm{a}_1,\ldots, \bm{a}_t\in \{0,1\}^{N}$, we  define a treatment history vector $\bar{\bm{a}}_t=(\bm{a}_0^\top,\bm{a}_1^\top,\ldots,\bm{a}_t^\top)^\top$ up to time $t$. For each $i\in \{1,\ldots,N\}$, let $S_{i,t+1}^*(\bar{\bm{a}}_{t}) \in \mathcal{S}_i$ and $R_{i,t}^*(\bar{\bm{a}}_t)\in \mathbb{R}$ be,  respectively,   the potential state and reward (outcome) associated with the $i$-th agent at time $t+1$ and time $t$, which  would occur had all agents followed $\bar{\bm{a}}_t$. 
Moreover,  different action histories would lead to different potential outcomes,  and more importantly, these potential outcomes cannot be directly observed.

We first introduce a  consistency assumption (CA) to link potential outcomes to the observed data.  
Let $\{(S_{i,t},A_{i,t},R_{i,t},S_{i,t+1})\}_{1\le i\le N,0\le t< T}$ be the observed data, where 
$(S_{i,t},A_{i,t},R_{i,t})$ stands for the observed state-action-reward triplet associated with the $i$-th agent at time $t$ and $T$ is the termination time of the study. 
Let $\bm{A}_t=(A_{1,t},\ldots,A_{N,t})^\top$  and $\bar{\bm{A}}_t=(\bm{A}_0^\top,\bm{A}_1^\top,\ldots,\bm{A}_t^\top)^\top$ 
be, respectively,  the observed treatments at time $t$ and until  time $t$.  The CA is given  as follows. 

\smallskip

\noindent (CA) $S_{i,t}=S_{i,t}^*(\bar{\bm{A}}_{t-1})$ and $R_{i,t}=R_{i,t}^*(\bar{\bm{A}}_t)$ hold almost surely for any $i$ and $t$.

\smallskip

Under CA, the potential outcomes are allowed to depend on not only past treatments, but also actions selected by other agents. 
The CA extends SUTVA to settings with spatiotemporal interference, since 
SUTVA requires $S_{i,t+1}^*$ and $R_{i,t}^*$ to be functions of $A_{i,t}$ only. {\color{black}We also remark that the notations $S_{i,t}^*$ and $R_{i,t}^*$ are used to denote potential outcomes. They are not deterministic functions of $\bar{\bm{A}}_{t-1}$ and $\bar{\bm{A}}_t$.}

We next introduce a sequential randomization assumption (SRA) that guarantees  the identifiability of our causal estimands. 

\smallskip

\noindent (SRA) $\bm{A}_t$ is independent of $\bm{W}^* $ given $\{(S_{i,j},A_{i,j},R_{i,j})\}_{1\le i\le N, 0\le j< t}\cup \{S_{i,t}\}_{1\le i\le N}$  for any $t$,  
\noindent where $\bm{W}^*=\cup_{t\ge 0,\bar{\bm{a}}_t\in \{0,1\}^{N(t+1)} } \bm{W}_t^*(\bar{\bm{a}}_t)$, in which   $\bm{W}_t^*(\bar{\bm{a}}_t)$ denotes the set of potential outcomes following $\bar{\bm{a}}_t$ up to time $t$, that is,
$\bm{W}_t^*(\bar{\bm{a}}_t)=\{ (S_{i,j}^*(\bar{\bm{a}}_{j-1}), R_{i,j}^*(\bar{\bm{a}}_j)):1\le i\le N, 0\le j\le t \}$.

\smallskip

{\color{black}We remark that SRA basically assumes that there is no unmeasured confounders. SRA is satisfied in randomized experiments, as in our real datasets. Under settings where SRA is violated, we can apply backdoor or frontdoor adjustment to the MARL setup to handle unmeasured confounders \citep[see e.g.,][]{wang2020provably}. This is beyond the scope of the current paper and we leave it for future research.} 
We also note that when $N=1$, i.e., in the single agent setting,  CA and SRA are commonly imposed in sequential decision making problems \citep[see e.g.,][]{Murphy2003,robins2004,zhang2013,ertefaie2014,Laber2018,luckett2019}. 

Next, we introduce the Markov assumption (MA) and the conditional mean independence assumption (CMIA) 
to characterize the system transitions. These assumptions serve as the foundations of the existing state-of-the-art RL algorithms \citep[see e.g.,][]{Sutton2018}. Let $\bm{S}_t=(S_{1,t},\cdots,S_{N,t})^\top$ and $\bm{R}_t=(R_{1,t},\cdots,R_{N,t})^\top$. 

\smallskip

\noindent (MA) There exists a Markov transition kernel $\mathcal{P}: \mathbb{S} \times \mathcal{A} \times \mathbb{S}$ such that for any $t \ge 0$, $\bar{\bm{a}}_t\in \{0,1\}^{N(t+1)}$ and  $\mathcal{S}\in \mathbb{S}$, we have almost surely that
\begin{eqnarray*}
	\prob \{\bm{S}_{t+1}\in \mathcal{S}|\bm{A}_t,\bm{S}_t,\{\bm{A}_{j},\bm{R}_j,\bm{S}_j\}_{0\le j<t}\}=\mathcal{P}(\mathcal{S};\bm{A}_{t},\bm{S}_{t}).
\end{eqnarray*} 

The transition kernel $\mathcal{P}$ characterizes the conditional distribution of the future state vector given the current state-action pair.
MA assumes the system dynamics are homogeneous over time. This enables consistent estimation of our causal estimands. We remark that MA is testable from the observed data; see e.g., the goodness-of-fit test developed by \cite{shi2020does}. 

\smallskip

\noindent (CMIA) There exist functions $r_1,\ldots,r_N$ such that for any $1\le i\le N$, $t\ge 0$, $\bar{\bm{a}}_t\in \{0,1\}^{N(t+1)}$, we have $\Mean (R_{i,t}|\bm{A}_t, \bm{S}_t,  \{\bm{A}_{j},\bm{R}_j,\bm{S}_j\}_{0\le j<t})=r_i(\bm{A}_t,\bm{S}_t)$ almost surely.

For each $1\le i\le N$, $r_i$ corresponds to the conditional mean function of $R_{i,t}$ given the state-action pair at time $t$. {\color{black}CMIA is similar to MA in the sense that it also imposes certain conditional independence assumption between the reward and the past data history. Meanwhile, it is weaker than requiring $R_{i,t}$ to be independent of $(A_j,R_j,S_j)_{0\le j<t}$ conditional on $(A_t,S_t)$, since it only requires the conditional mean of $R_{i,t}$ to satisfy the independence assumption.} It is also
weaker than the condition that requires the reward to be a deterministic function of the state-action pair. The latter condition is commonly imposed in the literature \citep[see e.g.,][]{ertefaie2014,luckett2019}. 

In practice, to ensure MA and CMIA to be satisfied, we can construct the state by concatenating measurements over multiple decision points till the Markov and conditional mean independence properties are satisfied. In addition, to guarantee the transition kernel $\mathcal{P}$ and the reward functions $\{r_i\}_i$ are time-homogeneous, we can include some auxiliary variables (e.g., time of the day) in the state. See our real data analysis in Section \ref{sec:real} for details. Without special saying, we assume that  CA, SRA, MA and CMIA hold throughout this paper. 

We  next
introduce the average treatment effect (ATE) for multi-agent systems below.  
We focus on the class of  stationary policies indexed by some $\bm{\pi}=(\pi_1,\ldots,\pi_N)^\top$,  where each $\pi_i$ is a binary-valued function of the current state vector. Under $\bm{\pi}$, the $i$-th spatial unit will receive the treatment $\pi_i(\bm{S}_t)$ at time $t$. As we have commented, applying a certain promotion strategy $\bm{\pi}$ in our application has both short-term and long-term benefits. 
We are thus interested in evaluating the average reward under $\bm{\pi}$. This allows the company to decide whether to apply such a dynamic policy in a given city or not, under some budget constraints. For any such policy $\bm{\pi}$, let $\bar{\bm{\pi}}_0=\bm{\pi}(\bm{S}_0)=(\pi_1(\bm{S}_0),\ldots,\pi_N(\bm{S}_0))^\top$  be the initial action vector assigned according to $\bm{\pi}$. Then we recursively define  $\bar{\bm{\pi}}_t=(\bar{\bm{\pi}}_{t-1}^\top,\bm{\pi}(S_{t}^*(\bar{\bm{\pi}}_{t-1})))^\top$ as the treatment assignment history under $\bm{\pi}$ up to time $t$ for $t\geq 1$.


{\color{black}Our objective is to evaluate the long term value of a given policy $\bm{\pi}$, defined as
\begin{eqnarray}\label{eqn:ATE}
	V(\bm{\pi})=\lim_{t\to \infty} \frac{1}{Nt}\sum_{i=1}^N \sum_{j=0}^t \Mean R_{i,j}^*(\bar{\bm{\pi}}_{j}),
\end{eqnarray}   
where  $R_{i,j}^*(\bar{\bm{\pi}}_{j})$
denotes the potential outcomes of the $i$-th agent that would occur at time $t$ had all agents followed the dynamic policy $\bm{\pi}$. 
We note that $V(\bm{\pi})$ can be represented by $N^{-1} \sum_{i=1}^N V_i(\bm{\pi})$,  where $V_i(\bm{\pi})=\lim_{t\to \infty} t^{-1}\sum_{j=0}^t \Mean R_{i,j}^*(\bar{\bm{\pi}}_j)$. Consequently, to evaluate $V(\bm{\pi})$, it suffices to estimate $V_i(\bm{\pi})$ for $i=1,\cdots,N$.} 

%
%


\section{Off-policy evaluation in MARL}\label{sec:method}
In this section,  we first propose an importance-sampling (IS) based estimator for $V_i(\bm{\pi})$ and then 
develop a doubly-robust version. We next detail some major steps in constructing these estimators. 




\subsection{IS based estimator.} 
In the following, we first consider a potential estimator for $V(\bm{\pi})$, which is built on the value estimator proposed by \cite{liu2018} in a single-agent system. We then discuss its limitation and present our IS based estimator. 

To detail the method, we assume the system follows a stationary behavior policy $b(\cdot)$  such that
$$\prob(\bm{A}_t=\bm{a}_t|\{A_{i,j},S_{i,j},R_{i,j} \}_{1\le i\le N, 0\le j<t}\cup \{S_{i,t}\}_{1\le i\le N} )=b(\bm{a}_t|\bm{S}_t),$$ holds  for any $\bm{a}_t\in \{0,1\}^N$. In other words, $\bm{A}_{t}$ depends on past observations only through $\bm{S}_{t}$. It implies that the process $\{(\bm{S}_t,\bm{A}_t)\}_{t\ge 0}$ forms a time-homogeneous Markov chain. This assumption is satisfied in our application where the data are generated from a completely randomized experiment with $b(\bm{A}_t|\bm{S}_t)=0.5^N$ for any $t$. Meanwhile, we also allow $b$ to rely on the set of current state variables.

Let $p_b(\bm{s})$ be the density function of the stationary distribution of the stochastic process $\{\bm{S}_t\}_{t\ge 0}$. 
Similarly, for a given $\bm{\pi}$, let $p_{\bm{\pi}}(\bm{s})$ be the stationary density function of $\{\bm{S}_{t}\}_{t\ge 0}$ had all agents followed $\bm{\pi}$.
When the process $\{\bm{S}_t\}_t$ reaches its stationary distribution, it follows from the change-of-measure theorem that
\vspace*{-0.1cm}
\begin{eqnarray}\label{eqn:valueliu2018}
\begin{split}
	V_i(\bm{\pi})=\int_{\mathbb{S}_i} \omega(\bm{s})r_i(\bm{\pi}(\bm{s}),\bm{s})p_b(\bm{s})d\bm{s}=\Mean \{\omega(\bm{S}_t) r_i(\bm{\pi}(\bm{S}_t),\bm{S}_t)\} 
	\\=\Mean\left\{ \omega(\bm{S}_t) \frac{\mathbb{I}(\bm{A}_t=\bm{\pi}(\bm{S}_t))}{b(\bm{\pi}(\bm{S}_t)|\bm{S}_t)} R_{i,t}\right\},
\end{split}
\end{eqnarray}
where $\omega(\bm{s})=p_{\bm{\pi}}(\bm{s})/p_b(\bm{s})$ and $\mathbb{I}(\cdot)$ denotes the indicator function. 
Thus, 
a  natural estimator for $V_i(\bm{\pi})$ is the  IS based estimator
$\widehat V_i^{IS_0}(\bm{\pi})=T^{-1}\sum_{t=0}^{T-1} \widehat{\omega}(\bm{S}_{t}) \mathbb{I}(\bm{A}_t=\bm{\pi}(\bm{S}_t))R_{i,t}/b(\bm{\pi}(\bm{S}_t)|\bm{S}_t)$
for some estimated $\widehat{\omega}$, leading to the IS estimator for $V(\bm{\pi})$: 
$$
\widehat V^{IS_0}(\bm{\pi})= (NT)^{-1} \sum_{i=1}^N \sum_{t=0}^{T-1} \frac{\widehat{\omega}(\bm{S}_{t}) \mathbb{I}(\bm{A}_t=\bm{\pi}(\bm{S}_t))R_{i,t}}{b(\bm{\pi}(\bm{S}_t)|\bm{S}_t)}. 
$$ 

In a multi-agent system, the above estimator  $\widehat V^{IS_0}_i(\bm{\pi})$ has two major limitations. The first one is that it suffers from   high variance introduced by the importance ratio $\omega(\bm{S}_{t}) \mathbb{I}(\bm{A}_t=\bm{\pi}(\bm{S}_t))/b(\bm{\pi}(\bm{S}_t)|\bm{S}_t)$. To better illustrate this, suppose that the state-action pairs are independent across different agents. In this case,  the overall ratio is the product of ratios associated with each single agent, so variances in each individual ratio accumulate multiplicatively. Thus,   the overall ratio can have an extremely high variance for large $N$. The second one  is that consistently estimation of $\omega(\cdot)$ is extremely challenging for high-dimensional state-action space and limited observations. One naive approach could  replace the overall weight in \eqref{eqn:valueliu2018} by the individual ratio associated with the $i$-th agent, but it would ignore the interference between different spatial units, leading to a biased value estimator.  

To address these two limitations, we propose to factorize the {\color{black}importance ratio} by using the mean-field approximation procedure. The key idea of this procedure is to approximate the reward function $r_i$ as a function of the state-action pairs of the $i$-th agent and its neighbours only. 
This allows us to focus on the density ratio of these restricted state-action pairs as detailed in \eqref{eqn:value}. As the input of such a density ratio is reduced to a moderate scale, the variance of the value estimator is dramatically reduced. It also enables consistent estimation of the density ratio.

We next detail the mean-field approximation procedure. For any $1\le i\le N$, let $\mathcal{N}(i)$ denote the index set of the neighboring agents of agent $i$. Let $m_{i}^s$ and $m_i^a$ be some mean-field functions of the local states and actions related to the $i$-th agent, respectively. For instance, one may  set these functions to some averaged state and action over its neighbors, i.e.,
\vspace*{-0.1cm}
\begin{eqnarray}\label{eqn:mf}
m_i^s(\bm{s})=\frac{1}{|\mathcal{N}(i)|}\sum_{j\in \mathcal{N}(i)} s_j\,\,\,\,\hbox{and} \,\,\,\,m_i^a(\bm{a})=\frac{1}{|\mathcal{N}(i)|}\sum_{j\in \mathcal{N}(i)} a_j,
\end{eqnarray}
for any $i$, where $|\mathcal{N}(i)|$ denotes the number of candidates in $\mathcal{N}(i)$ and $(s_i,a_i)$ corresponds to the state-action pair associated with the $i$-th agent. 
For each $i\in \{1,\ldots,N\}$ and any $\bm{s}\in \mathbb{S}$, $\bm{a}\in \{0,1\}^{N}$, we adopt the following mean-field approximation,
\begin{eqnarray}\label{eqn:MFA1}
r_i(\bm{a},\bm{s})= \bar{r}_i(a_i,m_i^a(\bm{a}),s_i,m_i^s(\bm{s})),
\end{eqnarray}
for some function $\bar{r}_i$. 


We make a few remarks. First, it is generally conceived that \eqref{eqn:MFA1} holds in many applications, such as the ride-sharing platform. Specifically, the state-action pair at one location can affect the outcome of other locations only through its impact on the distribution of drivers. Within each time unit, each driver can travel at most from one location to its neighbouring locations. Hence, the distribution of drivers in one location is independent of the state-action pairs in non-adjacent locations. 

{\color{black}Second, it is possible to test \eqref{eqn:MFA1} for some given mean-field functions $m_i^a$ and $m_i^s$. Specifically, notice that \eqref{eqn:MFA1} essentially requires the conditional mean of $R_{i,t}$ to be independent of $(\bm{A}_t,\bm{S}_t)$ given $(A_{i,t},m_i^a(\bm{A}_t),S_{i,t},m_i^s(\bm{S}_t))$. When $R_{i,t}$ satisfies the additive noise model assumption, $R_{i,t}=f(S_{i,t+1},A_{i,t},S_{i,t})+\varepsilon_{i,t}$, for some mean zero random error $\varepsilon_{i,t}$, \eqref{eqn:MFA1} holds if $S_{i,t+1}$ is conditionally independent of $(\bm{A}_t,\bm{S}_t)$ given $(A_{i,t},m_i^a(\bm{A}_t),S_{i,t},m_i^s(\bm{S}_t))$. As such, existing state-of-the-art conditional independence tests can be applied to test this assumption. We discuss this further in Section \ref{sec:numerical}.}

Recall that $p_b(\cdot)$ and $p_{\bm{\pi}}(\cdot)$ are the stationary distribution of $\bm{S}_t$ under $b$ and $\bm{\pi}$, respectively. Let $p_{i,\bm{\pi}}(\cdot)$ denote the corresponding marginal distribution of the triplet $\widetilde{S}_{i,t}=(m_{i}^{a}(\bm{\pi}(\bm{S}_t)),S_{i,t},m_{i}^s(\bm{S}_t))$. 
Similarly, we define  $p_{i,b}(\cdot)$.   
Let $\omega_i(\widetilde{S}_{i,t})$ denote the density ratio $p_{i,\bm{\pi}}(\widetilde{S}_{i,t})/p_{i,b}(\widetilde{S}_{i,t})$. It follows from similar arguments in \eqref{eqn:valueliu2018} that $V_i(\bm{\pi})$ equals 
\begin{eqnarray}\label{eqn:value}
\begin{split}
	&\int_{\tilde{s}_i} \omega_i(\tilde{s}_i)\bar{r}_i(\pi_i(\bm{s}),\tilde{s}_i)p_{i,b}(\tilde{s}_i)d\tilde{s}_i
	\\=&\Mean \omega_i(\widetilde{S}_{i,t})\frac{\mathbb{I}(A_{i,t}=\pi_i(\bm{S}_t),m_i^a(\bm{A}_t)=m_i^a(\bm{\pi}(\bm{S}_t)))}{b_i(\bm{\pi}|\widetilde{S}_{i,t})}R_{i,t},
\end{split}	
\end{eqnarray}
where $b_i(\bm{\pi}|\widetilde{S}_{i,t})$ denotes the conditional probability $\prob(A_{i,t}=\pi_i(\bm{S}_t),m_i^a(\bm{A}_t)=m_i^a(\bm{\pi}(\bm{S}_t))|\widetilde{S}_{i,t})$. In settings where $\bm{A}_t$ is independent of $\bm{S}_t$, as in our application, $b_i$ can be explicitly calculated. More generally, $b_i$ can be estimated by the state-of-the-art machine learning algorithms (see Appendix \ref{sec:propensity} in the supplement for details). 

Motivated by \eqref{eqn:value}, we consider a new IS based estimator of  $V_i(\bm{\pi})$ as
follows: 
\vspace*{-0.1cm}
\begin{eqnarray*}
\widehat{V}^{\scriptsize{\hbox{IS}}}_i(\bm{\pi})=\frac{1}{T} \sum_{t=0}^{T-1} \widehat{\omega}_i(\widetilde{S}_{i,t})\frac{\mathbb{I}(A_{i,t}=\pi_i(\bm{S}_t),m_i^a(\bm{A}_t)=m_i^a(\bm{\pi}(\bm{S}_t)))}{b_i(\bm{\pi}|\widetilde{S}_{i,t})}R_{i,t} 
\end{eqnarray*}
for some estimated $\widehat{\omega}_i$. Since the sampling ratio in $\widehat{V}^{\scriptsize{\hbox{IS}}}_i(\bm{\pi})$ is a function of $\widetilde{S}_{i,t}$, $A_{i,t}$ and $m_i^a(\bm{A}_t)$ only, $\widehat{V}^{\scriptsize{\hbox{IS}}}_i(\bm{\pi})$ has a much smaller variance compared to the value estimator outlined at the beginning of this section. In addition, consistent estimation of $\omega_i$ is feasible since  the dimension of the input of $\omega_i$ has been reduced to a moderate scale. Given $\widehat{V}_i^{\scriptsize{\hbox{IS}}}(\bm{\pi})$, the corresponding estimator for the average value $V(\bm{\pi})$ is given by $\widehat{V}^{\scriptsize{\hbox{IS}}}(\bm{\pi})=N^{-1} \sum_{i=1}^N \widehat{V}_i^{\scriptsize{\hbox{IS}}}(\bm{\pi})$. 

We discuss the estimating procedure for the density ratio $\omega_i$ in Section \ref{sec:weight}. 


%

\subsection{Doubly-robust estimator}
Compared to $\widehat{V}^{\scriptsize{\hbox{IS}}}(\bm{\pi})$, the doubly-robust (DR) estimator offers protection against model misspecification of the density ratio and is more efficient in general. \cite{kallus2019efficiently} developed a double reinforcement learning method for value evaluation in a single-agent discounted reward setting. In this section, we extend their proposal to a multi-agent  average reward setup.  
Before presenting the estimator, we introduce 
the Q-function associated with the $i$-th agent under a given policy $\bm{\pi}$ as
\begin{eqnarray*}
Q_i^{\bm{\pi}}(\bm{a},\bm{s})=\sum_{t=0}^{+\infty} \Mean[\{R_{i,t}^*(\bar{\bm{\pi}}_t(\bm{a}))-V_i(\bm{\pi})\}| \bm{S}_0=\bm{s} ],\,\,\,\,\forall \bm{s}\in \mathbb{S},\bm{a}\in \{0,1\}^N,
\end{eqnarray*}
where $\bar{\bm{\pi}}_t(\bm{a})$ denotes 
the treatment history up time $t$ such that the initial treatment equal to $\bm{a}$ and all other actions assigned according to $\bm{\pi}$. {\color{black}We remark that $Q_i^{\bm{\pi}}(\bm{a},\bm{s})$ is finite and well-defined when the Markov chain approaches its steady-state exponentially fast under $\bm{\pi}$. In that case, most of the differences on the right-hand-side will be earned in the first few iterations. Please see Section 8.2.1 of \cite{Puterman1994} for details.} 
%

The DR estimator for $V_i(\bm{\pi})$ takes the following form, 
\begin{eqnarray}\label{eqn:dr0}\\\nonumber
\widetilde{V}_i(\bm{\pi})+\frac{1}{T}\sum_{t=0}^{T-1} \widetilde{\omega}(\bm{S}_t) \frac{\mathbb{I}(\bm{A}_t=\bm{\pi}(\bm{S}_t))}{b(\bm{\pi}(\bm{S}_t)|\bm{S}_t)}\{R_{i,t}+\widetilde{Q}_i(\bm{\pi}(\bm{S}_{t+1}),\bm{S}_{t+1})-\widetilde{Q}_i(\bm{A}_t,\bm{S}_t)-\widetilde{V}_i(\bm{\pi})\},
\end{eqnarray}
where $\widetilde{V}_i(\bm{\pi})$ denotes some initial estimator for $V_i(\bm{\pi})$ and $\widetilde{\omega}$ and $\widetilde{Q}_i$ stand for estimators for $\omega$ and $Q_i$, respectively. Note that by the Bellman equation (see Lemma \ref{lemma:Q} in Section \ref{sec:Q} for details), the second term in \eqref{eqn:dr0} has zero mean when $(\widetilde{Q}_i,\widetilde{V}_i(\bm{\pi}))=(Q_i,V_i(\bm{\pi}))$. When $\widetilde{\omega}=\omega$, it can be shown that \eqref{eqn:dr0} has the same asymptotic mean as the IS-based estimator. 
Based on the above discussion, one can verify that \eqref{eqn:dr0} is consistent when either $\widetilde{\omega}=\omega$ or $(\widetilde{Q}_i,\widetilde{V}_i(\bm{\pi}))=(Q_i,V_i(\bm{\pi}))$. 

However, due to the presence of high-dimensional state-action space, the estimator outlined in \eqref{eqn:dr0} suffers from high variance. In addition, consistent estimation of $\omega$ and $Q_i$ are extremely difficult. To address these concerns, we replace the density ratio in (\ref{eqn:dr0}) by $\widehat{\omega}_i(\widetilde{S}_{i,t})\mathbb{I}(A_{i,t}=\pi_i(\bm{S}_t),m_i^a(\bm{A}_t)=m_i^a(\bm{\pi}(\bm{S}_t)))/b_i(\bm{\pi}|\widetilde{S}_{i,t})$. 
To enable consistent estimation of $Q_i$, we consider factorizing $Q_i$ based on mean-field approximation as well. 
Specifically, for each $i\in \{1,\ldots,N\}$, 
and any $s\in \mathbb{S}$, $\bm{a}\in \{0,1\}^N$, we propose to approximate $Q_i$ by
\begin{eqnarray}\label{eqn:MFA2}
Q_i^{\bm{\pi}}(\bm{a},\bm{s})=\bar{Q}_i(a_i,m_i^a(\bm{a}),s_i,m_i^s(\bm{s})),
\end{eqnarray} 
for some functions $\bar{Q}_i$. 

{\color{black}When \eqref{eqn:MFA1} holds, we show that \eqref{eqn:MFA2} is satisfied if $(\pi(\bm{S}_{t+1}),m_i^a(\bm{\pi}(\bm{S_{t+1}})),\widetilde{S}_{i,t+1})$ is conditionally independent of $(\bm{A}_t,\bm{S}_t)$ given $(A_{i,t},m_i^a(\bm{A}_t),S_{i,t},m_i^s(\bm{S}_t))$. See Appendix \ref{app:moremf} for details. Similarly, existing state-of-the-art conditional independence tests can be applied to verify this assumption.}

To learn $\bar{Q}_i$ and $V_i(\bm{\pi})$, we extend the regularized policy iteration algorithm \citep{Farah2016,liao2019off} to our setup. The detailed procedure is given in Section \ref{sec:Q}. Meanwhile, other methods developed in single-agent systems \citep[see e.g.,][]{uehara2019minimax} may be adopted as well.   
Let $\widehat{Q}_i$ and $\widehat{V}_i(\bm{\pi})$ denote the corresponding estimators, we define our value estimator 
\begin{eqnarray}\label{widehatViDR}
\begin{split}
	\widehat{V}^{\hbox{\scriptsize{DR}}}_i(\bm{\pi})=\widehat{V}_i(\bm{\pi})+\frac{1}{T}\sum_{t=0}^{T-1} \widehat{\omega}_{i}(\widetilde{S}_{i,t})\frac{\mathbb{I}(A_{i,t}=\pi_i(\bm{S}_t),m_i^a(\bm{A}_t)=m_i^a(\bm{\pi}(\bm{S}_t)))}{b_i(\bm{\pi}|\widetilde{S}_{i,t})}\\
	\times \{R_{i,t}+\widehat{Q}_i(\pi_i(\bm{S}_{t+1}),\widetilde{S}_{i,t+1})-\widehat{Q}_i(A_{i,t},m_i^a(\bm{A}_t),S_{i,t},m_i^s(\bm{S}_t))-\widehat{V}_i(\bm{\pi})\}.
\end{split}	
\end{eqnarray}
The corresponding estimator for $V(\bm{\pi})$ is given by $\widehat{V}^{\hbox{\scriptsize{DR}}}(\bm{\pi})=N^{-1} \sum_{i=1}^N \widehat{V}^{\hbox{\scriptsize{DR}}}_i(\bm{\pi})$. 

To conclude this section, we present an overview of our theoretical results for $\widehat{V}^{\hbox{\scriptsize{DR}}}(\bm{\pi})$. Details are given in the appendix. 
Our theoretical studies are mostly concerned with an ``oracle" estimator $\widehat{V}^{\hbox{\scriptsize{DR}}*}(\bm{\pi})$,  which works as if the true values $\bar{Q}_i$, $\omega_i^*$ and $V_i^*(\bm{\pi})$ were known. Specifically, let $\widehat{V}_i^{\hbox{\scriptsize{DR}}*}(\bm{\pi})$ be a version of $\widehat{V}^{\hbox{\scriptsize{DR}}}(\bm{\pi})$ by replacing $\widehat{\omega}_i$, $\widehat{Q}_i$ and $\widehat{V}_i(\bm{\pi})$ in \eqref{widehatViDR} with the corresponding population limits.
%
The oracle estimator is given by $\widehat{V}^{\hbox{\scriptsize{DR}}*}(\bm{\pi})=N^{-1}\sum_{i=1}^N \widehat{V}_i^{\hbox{\scriptsize{DR}}*}(\bm{\pi})$. 
In Theorem \ref{thm:oracle}, we establish the doubly-robustness property of the oracle estimator. Specifically, we show the oracle estimator is $(NT)^{-1/2}$-consistent and asymptotically normal when one of the mean-field approximation is valid, i.e., either \eqref{eqn:MFA1} or \eqref{eqn:MFA2} holds. In Theorem \ref{thm:double},  we establish the doubly-robustness property of our estimator, i.e., $\widehat{V}^{\hbox{\scriptsize{DR}}}(\bm{\pi})$ is consistent when either \eqref{eqn:MFA1} or \eqref{eqn:MFA2} holds. In Theorem \ref{thm:oracleest}, we show our value estimator achieves the ``oracle" property when both mean-field approximations  \eqref{eqn:MFA1} and \eqref{eqn:MFA2} are valid. Specifically, it is $(NT)^{-1/2}$-consistent and asymptotically normal with the asymptotic variance equal to that of the oracle estimator. {\color{black}We remark that to establish these theoretical results, we require the state-action process $\{(\bm{S}_t,\bm{A}_t)\}$ to satisfy the exponential $\beta$-mixing condition; see (A1) in Appendix \ref{sec:theory}. Under stationarity, this assumption is equivalent to require the underlying Markov chain to satisfy geometric ergodicity \citep[see Theorem 3.7 of][]{Bradley2005}. It guarantees that the estimated values concentrates on their oracle values with high probability, allowing us to establish the oracle property. Please refer to the proof of Theorem \ref{thm:oracleest} in Appendix \ref{app:proofthmoracleest} for details.}
\subsection{Estimation of the density ratio}\label{sec:weight}
In the single-agent setup, there are multiple estimation methods available to produce an estimated density ratio \citep{liu2018,kallus2019efficiently,nachum2019dualdice,liao2020batch}. In our implementation, we extend the proposal in \cite{liu2018} to the multi-agent setup. We first present an overview of the algorithm. A key observation is given by Lemma \ref{lemma:weight}, which establishes the relationship between $\omega_i(\widetilde{S}_{i,t})$ and $\omega_i(\widetilde{S}_{i,t+1})$. Based on this lemma, 
the idea is to 
introduce a discriminator function to construct a mini-max loss function (see Equation \eqref{eqn:omegai}). Then $\omega_i$ is estimated by optimizing this loss function. We next present Lemma \ref{lemma:weight}.

\begin{lemma}\label{lemma:weight}
Suppose $\widetilde{S}_{i,t+1}$ is independent of $(\bm{S}_t,\bm{A}_t)$ given $\widetilde{S}_{i,t}$, $A_{i,t}$ and $m_i^a(\bm{A}_t)$. We have $\Mean \Delta_{i,t}(\omega_i) f(\widetilde{S}_{i,t+1})=0$ for any $i,t$ and function $f$ where
\vspace*{-0.1cm}
\begin{eqnarray*}
	\Delta_{i,t}(\omega_i)=\omega_i(\widetilde{S}_{i,t})\frac{\mathbb{I}(A_{i,t}=\pi_i(\bm{S}_t),m_i^a(\bm{A}_t)=m_i^a(\bm{\pi}(\bm{S}_t)))}{b_i(\bm{\pi}|\widetilde{S}_{i,t})}-\omega_i(\widetilde{S}_{i,t+1}).
\end{eqnarray*}
\end{lemma}

Under the conditions in Lemma \ref{lemma:weight}, $\{(\widetilde{S}_{i,t},A_{i,t},m_i^a(\bm{A}_t))\}_{t\ge 0}$ forms a time-homogeneous Markov chain. This lemma motivates us to compute $\widehat{\omega}_i$ by minimizing the following loss function,
\begin{eqnarray}\label{eqn:omegai}
\widehat{\omega}_i=\argmin_{\omega_i\in \Omega} \sup_{f\in \mathcal{F}} \left|\sum_{t=0}^{T-1} \Delta_{i,t}(\omega_i) f(\widetilde{S}_{i,t+1})\right|^2,
\end{eqnarray}
for some function classes $\Omega$ and $\mathcal{F}$. In our implementation, we set $\mathcal{F}$ to a unit ball of a reproducing kernel Hilbert space (RKHS), i.e., 
\begin{eqnarray*}
\mathcal{F}=\{f\in \mathcal{H}:\|f\|_{\mathcal{H}}=1\},
\end{eqnarray*}
where 
\begin{eqnarray*}
\mathcal{H}=\left\{f(\cdot)=\sum_{t=0}^{T-1} b_t \kappa(\widetilde{S}_{i,t+1};\cdot): \{b_t\}_{t=0}^{T-1}\in \mathbb{R}^{T} \right\},
\end{eqnarray*}
for some positive definite kernel $\kappa(\cdot;\cdot)$ and $\|\cdot\|_{\mathcal{H}}$ denotes the corresponding RKHS norm. 

\begin{figure}[!t]
\includegraphics[width=6cm]{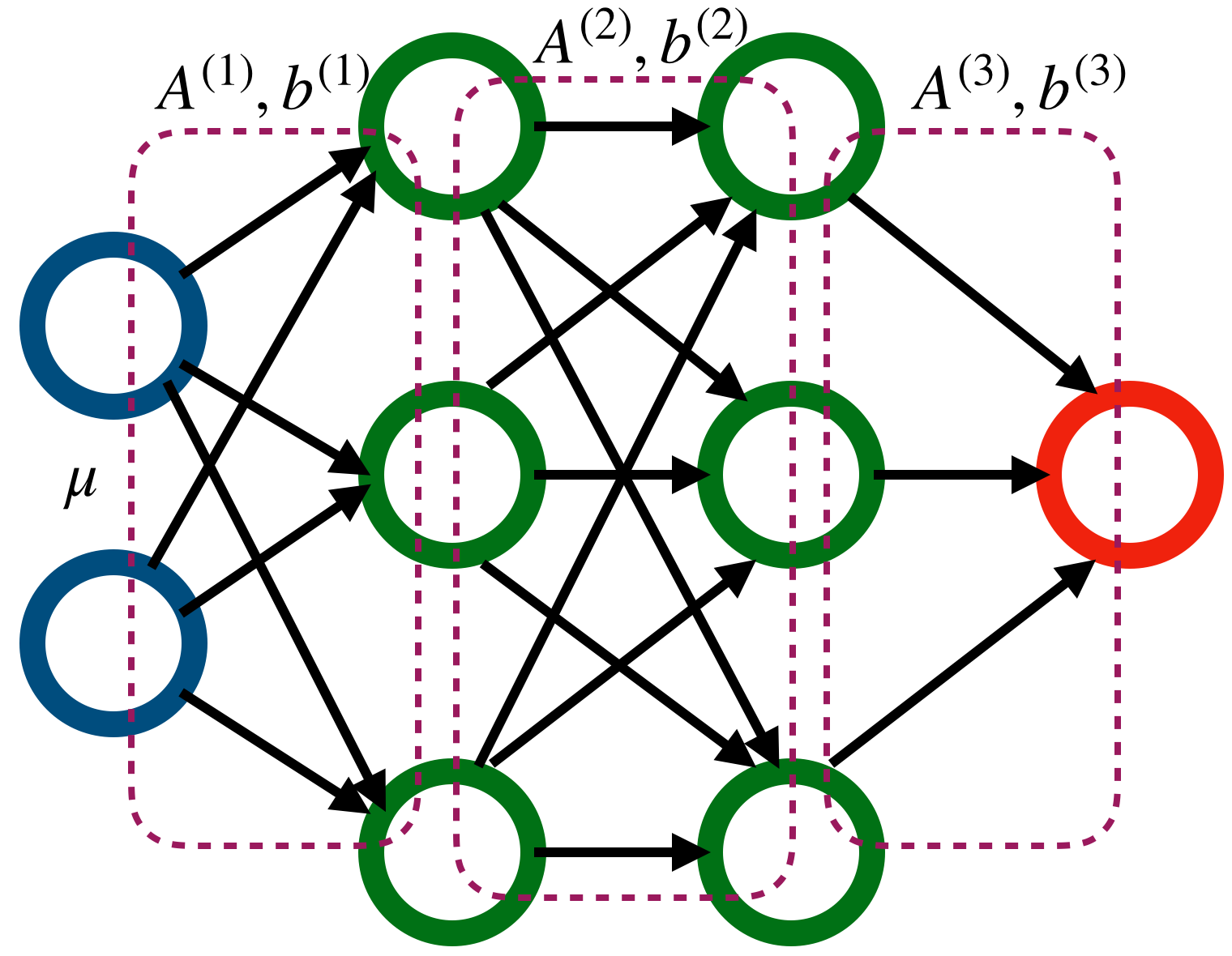}
\centering
\caption{Illustration of multilayer perceptron with two hidden layers, $m_0=2$, $m_1=m_2=3$. Here $u$ is the input, $A^{(\ell)}$ and $b^{(\ell)}$ denote the corresponding parameters to produce the linear transformation for the $(\ell-1)$th layer.}\label{fig:MLP}
\end{figure}

The use of RKHS enables us to derive a close-form expression for the objective function on the right-hand-side (RHS) of \eqref{eqn:omegai}. Specifically, using similar arguments in the proof of Theorem 2 of \cite{liu2018}, the optimization problem in \eqref{eqn:omegai} is then reduced to 
\begin{eqnarray*}
\widehat{\omega}_i=\argmin_{\omega_i\in \Omega} \sum_{t_1=0}^{T-1} \sum_{t_2=0}^{T-1} \Delta_{i,t_1}(\omega_i)\Delta_{i,t_2}(\omega_i) \kappa(\widetilde{S}_{i,t_1+1},\widetilde{S}_{i,t_2+1}). 
\end{eqnarray*}
It remains to specify the function class for $\Omega$. Motivated by the approximation capabilities of neural networks, we set $\Omega$ to the class of multilayer perceptron networks. See Figure \ref{fig:MLP} for an illustration. We use different parameters to factorize different $\omega_i$ such that each $\widehat{\omega}_i$ is computed separately. Alternatively, one could allow different $\omega_i$ to share some common parameters. Stochastic gradient descent is applied to update the parameters in the neural network. We detail our procedure in Algorithm \ref{alg1}. 

\begin{algorithm}[t!]
\caption{Estimation of the density ratio.}
\label{alg1}
\begin{algorithmic}
	\item
	\begin{description}
		\item[\textbf{Input}:] The data $\{(S_{i,j},A_{i,j},R_{i,j}):1\le i\le N,0\le j< T\}$. A target policy $\bm{\pi}$. 
		
		\item[{\color{black}\textbf{Initialize}}:] Initial the density ratio $\omega_i=\omega_{i,\theta}$ for $1\le i\le N$, to be  some neural networks parameterized by $\theta$.
		
		\item[\textbf{for}] iteration $=1,2,\cdots$ \textbf{do}
		\begin{enumerate}
			\item[a] Randomly sample a batch $\mathcal{M}$ from $\{0,1,\cdots,T-1\}$.
			
			\item[b] {\textbf{Update}} the parameter $\theta$ by $\theta\leftarrow \theta-\epsilon N^{-1}\sum_{i=1}^N \nabla_{\theta} D_i(\omega_{i,\theta}/z_{\omega_{i,\theta}})$ where $D_i(\omega_{i,\theta})$ is equal to
			\begin{eqnarray*}
				\frac{1}{|\mathcal{M}|}\sum_{t_1,t_2\in \mathcal{M}}  \Delta_{i,t_1}(\omega_{i,\theta})\Delta_{i,t_2}(\omega_{i,\theta}) \kappa(\widetilde{S}_{i,t_1+1},\widetilde{S}_{i,t_2+1}),
			\end{eqnarray*}
			and $z_{\omega_{i,\theta}}$ is a normalization constant $z_{\omega_{i,\theta}}=|\mathcal{M}|^{-1} \sum_{t\in \mathcal{M}} \omega_{i,\theta}(\widetilde{S}_{i,t+1})$. 
		\end{enumerate}
		\item[\textbf{Output}] $\omega_{i,\theta}$ for $1\le i\le N$. 
	\end{description}
\end{algorithmic}
\end{algorithm}

\subsection{Estimation of the Q-function and the value}\label{sec:Q}
In this section, we describe methods to estimate compute $\widehat{Q}_i$ and the initial value estimator $\widehat{V}_i(\bm{\pi})$. The key ingredient of the algorithm lies in minimizing a regularized version of the Bellman residual to work with rich nonparametric function class, while simultaneously controlling  its complexity. In our implementation, we use the RKHS as the function class to approximate the Q-function. 
A key observation is given by the following lemma,

\begin{lemma}\label{lemma:Q}
$\Mean \{R_{i,t}+	Q_i^{\bm{\pi}}(\bm{\pi}(\bm{S}_{t+1}),\bm{S}_{t+1})|\bm{S}_t,\bm{A}_t\}=V_i(\bm{\pi})+Q_i^{\bm{\pi}}(\bm{A}_t,\bm{S}_t)$ almost surely for any $i,t$. 
\end{lemma}

The equation in Lemma \ref{lemma:Q} is well-known as the Bellman equation in the average-reward case. Based on Lemma \ref{lemma:Q}, it is tempting to estimate $Q_i^{\bm{\pi}}$ by directly minimizing 
\begin{eqnarray}\label{eqn:obj}
\argmin_{(V_i,Q_i)}T^{-1} \sum_{t}  \{R_{i,t}+	Q_i(\bm{\pi}(\bm{S}_{t+1}),\bm{S}_{t+1})-V_i-Q_i(\bm{A}_t,\bm{S}_t)\}^2.
\end{eqnarray}
However, the resulting estimators are known to be biased when the Markov transition function is not deterministic \citep[see e.g.,][]{Farah2016}. To elaborate this, we note that the population limit of the objective function in \eqref{eqn:obj} equals 
\begin{eqnarray}\label{eqn:obj0}
\begin{split}	
	T^{-1}\sum_{t}  \Mean [\Mean \{R_{i,t}+	Q_i(\bm{\pi}(\bm{S}_{t+1}),\bm{S}_{t+1})|\bm{S}_t,\bm{A}_t\}-V_i-Q_i(\bm{A}_t,\bm{S}_t)]^2\\
	+T^{-1}\sum_{t} \Mean [\Var \{R_{i,t}+	Q_i(\bm{\pi}(\bm{S}_{t+1}),\bm{S}_{t+1})|\bm{S}_t,\bm{A}_t\}].
\end{split}	
\end{eqnarray}
The first line equals zero when $Q_i=Q_i^{\bm{\pi}}$ and $V_i=V_i(\bm{\pi})$. However, the second line depends on $Q_i$ as well. As such, $(Q_i^{\bm{\pi}},V_i(\bm{\pi}))$ might not necessarily be the minimizer of \eqref{eqn:obj0}. 

To resolve this issue, we consider first estimating the residual $\Mean \{R_{i,t}+	Q_i(\bm{\pi}(\bm{S}_{t+1}),\bm{S}_{t+1})|\bm{S}_t,\bm{A}_t\}\\-V_i-Q_i(\bm{A}_t,\bm{S}_t)$ as a function of $\bm{A}_t$, $\bm{S}_t$, $Q_i$, $V_i$ and then minimizing a regularized version of the squared residual. Under \eqref{eqn:MFA2}, we factorize $Q_i(\bm{a},\bm{s})$ by $Q_i(a_i,m_i^a(\bm{a}),s_i,m_i^s(\bm{s}))$. This yields the following optimization, 
\begin{eqnarray}\label{eqn:obj1}
(\widehat{V}_i(\bm{\pi}),\widehat{Q}_i)=\argmin_{(\eta,Q_i)\in \mathbb{R}\times \mathcal{Q}}\frac{1}{T}\sum_{t=0}^{T-1} \widehat{g}_{i}^2(A_{i,t},m_i^a(\bm{A}_t),S_{i,t},m_i^s(\bm{S}_t);\eta,Q_i)+\lambda \|Q_i\|_{\mathcal{Q}}^2,
\end{eqnarray}
where 
\begin{eqnarray}\label{eqn:gQ}
\begin{split}
	&\widehat{g}_{i}(\cdot,\cdot,\cdot,\cdot,\cdot;\eta,Q_i)=\argmin_{g\in \mathcal{G}}\frac{1}{T}\sum_{t=0}^{T-1} \{R_{i,t}+Q_{i}(\pi_i(\bm{S}_{t+1}),\widetilde{S}_{i,t+1})-\eta\\
	-&Q_i(A_{i,t},m_i^a(\bm{A}_t),S_{i,t},m_i^s(\bm{S}_t))-g(A_{i,t},m_i^a(\bm{A}_t),S_{i,t},m_i^s(\bm{S}_t))\}^2+\mu \|g\|_{\mathcal{G}}^2,
\end{split}	
\end{eqnarray}
and $\mu$ and $\lambda$ stand for some tuning parameters and $\|\cdot\|_{\mathcal{Q}}$, $\|\cdot\|_{\mathcal{G}}$ denote the corresponding RKHS norms. {\color{black}The purpose of adding the regularization terms in \eqref{eqn:obj1} and \eqref{eqn:gQ} is to prevent overfitting and guarantee the consistency of the estimator Q-function. Specifically, without the regularization term in \eqref{eqn:obj1}, the estimated Q-function overfits to the noise and becomes inconsistent. The regularization term in \eqref{eqn:gQ} guarantees that the population limit of the RHS of \eqref{eqn:obj1} converges to the first line of \eqref{eqn:obj0}.
Without this regularization term, the coupled optimization problems reduces to \eqref{eqn:obj}, yielding a biased solution. }

Next we derive the close-form expressions of 	$(\widehat{V}_i(\bm{\pi}),\widehat{Q}_i)$. Let $Z_{i,t}=(A_{i,t},m_i^a(\bm{A}_t),S_{i,t},m_i^s(\bm{S}_t))^\top$ and $Z_{i,t}^*=(\pi_i(\bm{S}_{t+1}),\widetilde{S}_{i,t+1})^\top$. Let $K_g$ and $K_Q$ denote the reproducing kernels used to model $g$ and $Q$, respectively. In our implementation, we use Gaussian RBF kernels to model these two functions. For a given $Q_i$ and $\eta$, the optimizer of \eqref{eqn:obj0} $\widehat{g}_i$ can be represented by $\sum_{t=0}^{T-1} \widehat{\beta}_{i,t} K_g(Z_{i,t},\cdot)$. With some calculations, we obtain $\widehat{\bm{\beta}}_i=(\widehat{\beta}_{i,0},\cdots,\widehat{\beta}_{i,T-1})^\top$ as
\begin{eqnarray*}
\widehat{\bm{\beta}}_i=\argmin_{\bm{\beta}} \frac{1}{T}\sum_{t=0}^{T-1} \left\{R_{i,t}+Q_i(Z_{i,t}^*)-\eta-Q_i(Z_{i,t})-\sum_{j=0}^{T-1} \beta_{j} K_g(Z_{i,j},Z_{i,t}) \right\}^2+\mu \bm{\beta}^\top \bm{K}_g \bm{\beta}\\
=\frac{1}{T} \bm{\beta}^\top \{\bm{K}_g\bm{K}_g^\top+T\mu \bm{K}_g\} \bm{\beta}-\frac{2}{T}\bm{\beta}^\top \bm{K}_g (\bm{R}+\bm{Q}_i^{*}-\bm{Q}_i-\eta \bm{1})\\
+\hbox{some~terms~that~are~independent~of~}\bm{\beta},
\end{eqnarray*}
where $\bm{K}_g=\{K_g(Z_{i,j_1},Z_{i,j_2})\}_{j_1,j_2}$ and $\bm{R}$, $\bm{Q}_i^{*}$ and $\bm{Q}_i$ are the column vectors formed by elements in $R_t$, $Q_i(Z_{i,t}^*)$ and $Q_i(Z_{i,t})$, respectively. This allows us to derive a close-form expression for $\widehat{\bm{\beta}}_i$. See Appendix \ref{sec:propensity} for details. As a result, for a given $Q_i$ and $\eta$, we have
\begin{eqnarray*}
\widehat{g}_i(Z_{i,t};\eta,Q_i)=\widehat{\bm{\beta}}_i^\top \bm{K}_g \bm{e}_t,
\end{eqnarray*}
where $\bm{e}_t$ denotes the column vector with the $t$-th element equals to one and other elements equal to zero. As such,
\begin{eqnarray*}
\frac{1}{T}\sum_{t=0}^{T-1} \widehat{g}_{i}^2(A_{i,t},m_i^a(\bm{A}_t),S_{i,t},m_i^s(\bm{S}_t);\eta,Q_i)=\frac{1}{T}\widehat{\bm{\beta}}_i^\top \bm{K}_g \bm{K}_g^T \widehat{\bm{\beta}}_i.
\end{eqnarray*}
Similarly, we can represent $Q_i$ as $\sum_{t=0}^{2T-1} \widehat{\alpha}_{i,t} K_Q(\widetilde{Z}_{i,t},\cdot)$ where $\widetilde{Z}_{i,t}$ denotes the $t$-th element in the vector $(Z_{i,0}^\top,\cdots,Z_{i,T-1}^\top,Z_{i,0}^{*\top},\cdots,Z_{i,T-1}^{*\top})^\top$. The closed-form expression of $\widehat{\bm{\alpha}}_i=(\widehat{\alpha}_{i,0},\cdots,\widehat{\alpha}_{i,T-1})^\top$ can be similarly obtained. Details are given in Appendix \ref{sec:propensity}. 

\section{Simulations}\label{sec:numerical}

In this section, we design an example to simulate the environment of a ride-sharing platform. Specifically, we consider orders and drivers operating in a map of $5\times 5$ spatial grids. For each grid, we design three time-varying variables to construct the state. Let $D_{i,t}$ and $O_{i,t}$, respectively,  denote the number of drivers and orders in the $i$-th grid during the time interval $(t-1,t]$. In practice, these two factors are known to have large impact on the driver income and customer satisfaction. The last variable $M_{i,t}$ measures the degree of mismatch between orders and drivers in the $i$-th grid at time $t$. Specifically, we set $M_{i, t} = 0.5 \{1-|D_{i,t}-O_{i,t}|/(1 + D_{i,t}+O_{i,t})\} + 0.5M_{i,t-1}$. Given the state vector at time $t+1$, we generate the reward $R_{i,t}$ from the following model: \begin{eqnarray}\label{eqn:addnoise}
R_{i,t} = M_{i, t + 1}\min(D_{i, t + 1}, O_{i, t + 1}) + \varepsilon_{i,t},
\end{eqnarray} 
where $\{\varepsilon_{i,t}\}_{i,t}$ are i.i.d. $\mathcal{N}(0,\sigma^2_R)$. At each time $t$, the platform can decide whether to implement a certain driver-side subsidizing policy to the $i$th spatial unit or not. This yields a binary action $A_{i,t}$. In our experiment, $\{A_{i,t}\}_{i,t}$ are i.i.d. according to a Bernoulli distribution with success probability $0.5$. 

Orders are simulated in the following manner. For $i=1,\ldots,25$, we first randomly generate $\mu_i$ from $N(100,25^2)$. Then we independently generate $O_{i,t}$'s from a Poisson distribution with expectation $\mu_i$. Thus, each $\mu_i$ represents the average number of orders during each time unit in region $i$. We plot these $\mu_i$'s in Figure \ref{fig11} (a). 

Drivers are simulated in the following manner. Initially, we put 80 drivers in each grid. At each time, drivers will be attracted to nearby regions that either implement the subsidizing policy or has large number of orders. To characterize this effect, at each time, we assign an attraction rate parameter to each region as $\nu_{i,t} = 1.5\exp(A_{i,t}) + 0.5(O_{i,t}/D_{i,t})$. At time $t+1$, drivers are more likely to move to neighborhood regions with large attraction rates. Specifically, we set $D_{i, t + 1} = \nu_{i,t}D_{i,t} (\sum_{j \in \mathcal{N}(i) }\nu_{j,t})^{-1}$.

\begin{figure}[!t]
\centering
\includegraphics[width=0.8\textwidth]{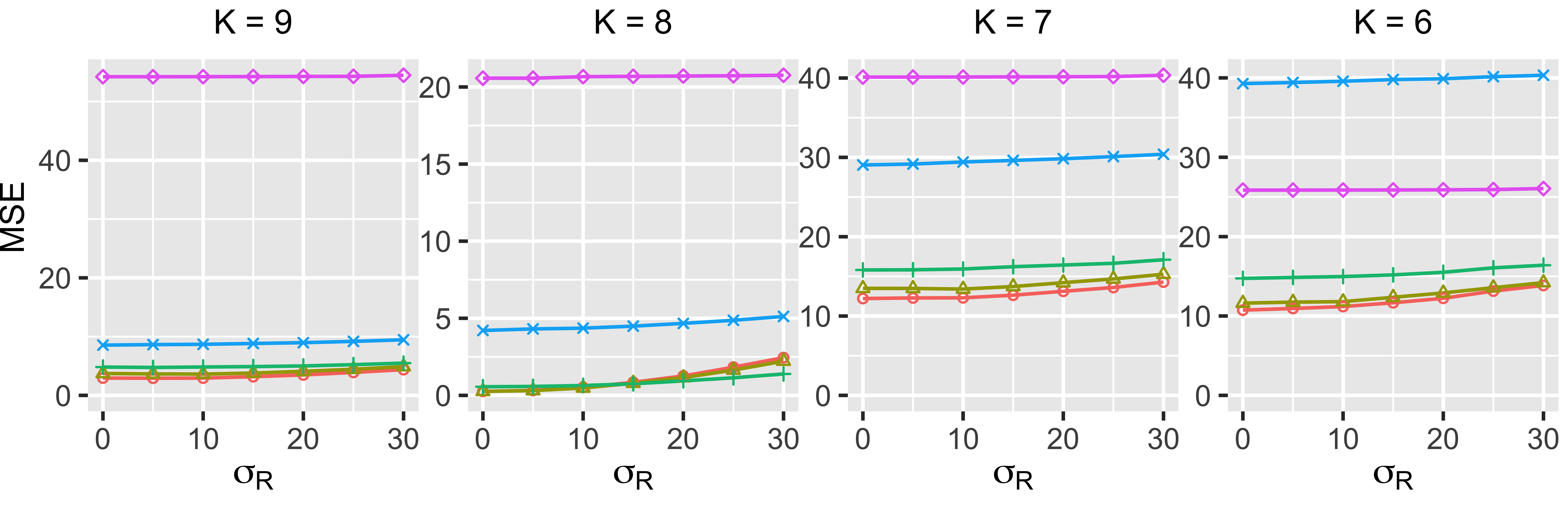} \\
\includegraphics[width=0.8\textwidth]{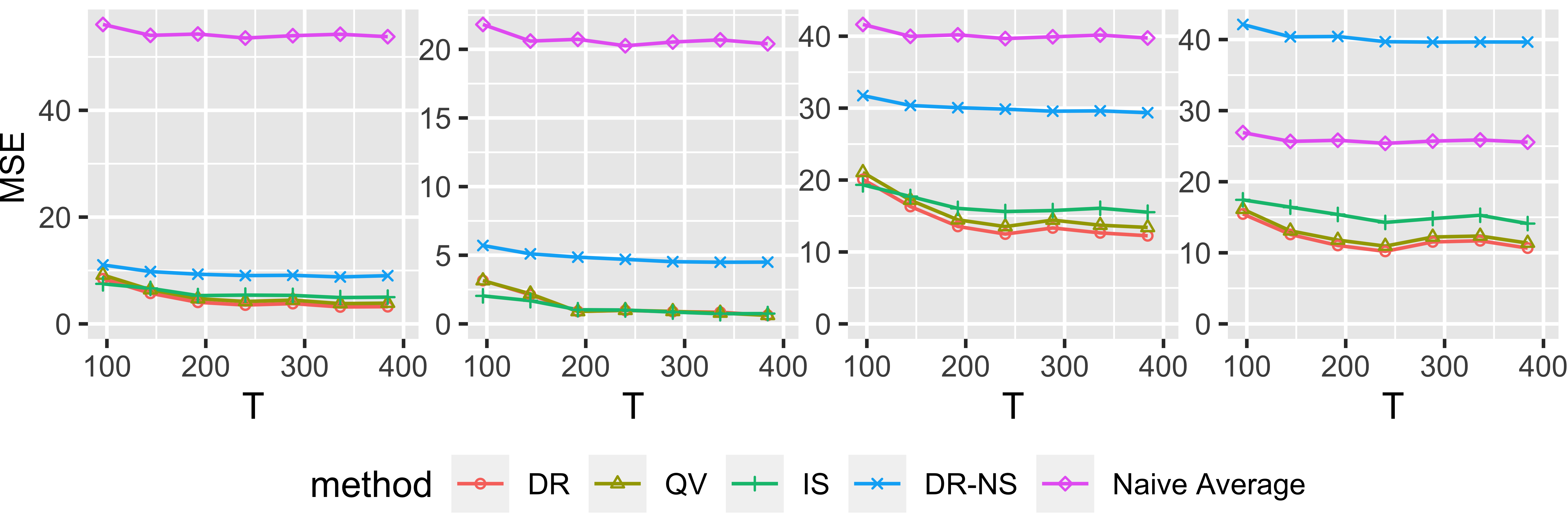} \\
\caption{Mean squared errors of different value estimates, aggregated over 100 simulations. $T$ is set to $336$ (the experiment lasts for two weeks and each hour is treated as one time unit) in the top plots and $\sigma_R$ is set to $15$ in the bottom plots. {\color{black}MSEs of DR-NM are larger than $1000$ for all choices of $\sigma_R$ and $T$ and  thus are  not plotted.}}
\label{figure:simu}
\end{figure}

Under the current setup, to increase the long-term reward, we can implement subsidizing policies in regions with large $\mu_i$'s. Specifically, we focus on four nondynamic policies $\{\bm{\pi}_K\}_{K \in \{6,7,8,9\}}$. Under $\bm{\pi}_K$, the subsidizing policies will be implemented in the top $K$ regions with largest $\mu_i$'s at each time. We are interested in evaluating the values $V(\bm{\pi}_K)$ for $K=6,7,8,9$. {\color{black}The true value of $V(\bm{\pi}_K)$ can be approximated via Monte Carlo simulations. Specifically, in each simulation, we generate data trajectories $\{(S_{i,t},A_{i,t},R_{i,t})\}_{1\le i\le N, 0\le t<M}$ under the target policy $\bm{\pi}_K$ for some sufficiently large integer $M>0$, and calculate the average reward $(NM)^{-1}\sum_{i=1}^N\sum_{t=0}^{M-1}R_{i,t}$. Then we aggregate these rewards over hundreds of simulations and treat it as the true value of $V(\bm{\pi})$. These estimated true values ranges from 55 to 60, with standard errors ranges from 0.05 to 0.35, across different settings}. 
Comparisons are made among the following methods: \\
(a) The doubly-robust estimator (denoted by DR) proposed in Section \ref{sec:method}; \\
(b) The IS based estimator (denoted by IS) proposed in Section \ref{sec:method}; \\
(c) The doubly-robust estimator without considering the spatial dependence (denoted by DR-NS); \\
(d) The doubly-robust estimator in (\ref{eqn:dr0}) without the mean field approximation (denoted by DR-NM);\\
{\color{black}(e) The Q-function based value estimator $N^{-1}\sum_{i=1}^N \widehat{V}_i(\bm{\pi})$ where each $\widehat{V}_i(\bm{\pi})$ is computed according to Section \ref{sec:Q} (denoted by QV); }\\
(f) A naive average of all immediate rewards.

\begin{figure}[!t]
\includegraphics[width=5cm]{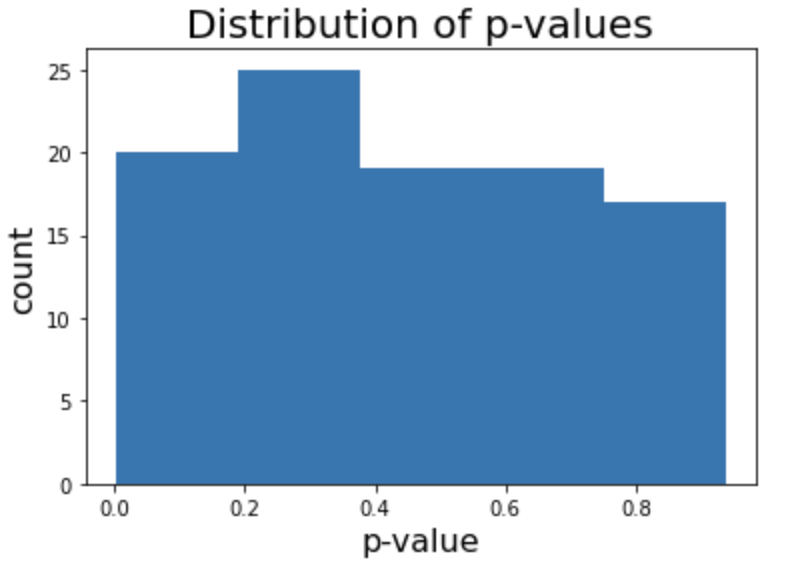}
\includegraphics[width=5cm]{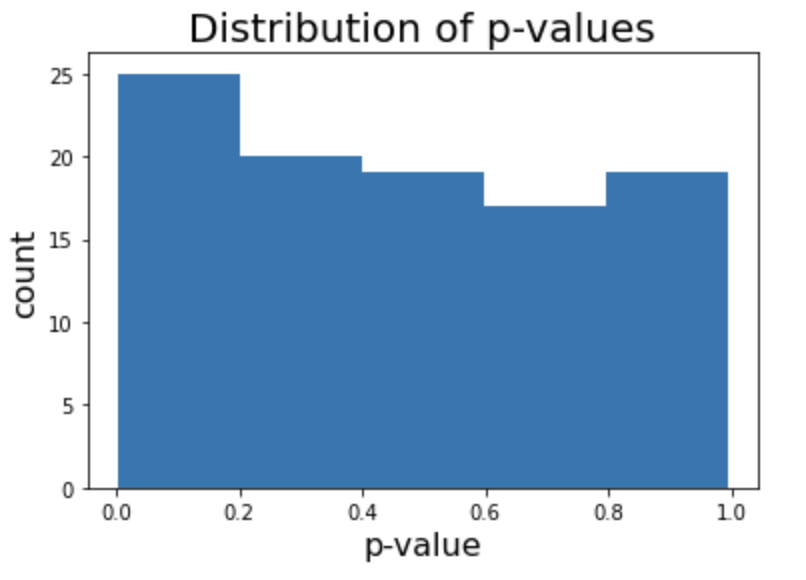}
\caption{Distribution of p-values for testing the mean-field approximation assumption associated with two randomly selected regions.}\label{fig:mf}
\end{figure}

{\color{black}To implement the proposed estimator, we set the mean-field functions to the averaged state and action over the neighbors, as in \eqref{eqn:mf}. Before presenting the results, we first investigate the validity of the mean-field approximation assumption with such a choice of mean-field functions. In our simulations, the reward is generated according to the additive noise model (see \eqref{eqn:addnoise}). As we have commented, it suffices to test the conditional independence of $(\pi_i(\bm{S}_{t+1}),m_i^a(\bm{\pi}(\bm{S}_{t+1})),\widetilde{S}_{i,t+1})$ and $(\bm{A}_t,\bm{S}_t)$ given $(A_{i,t},m_i^a(\bm{A}_t),S_{i,t},m_i^s(\bm{S}_t))$. Notice that according to the definition of $\bm{\pi}_K$, the evaluation policies depend only on $\mu_i$'s (which are fixed over different simulations) and are thus state-agnostic, e.g., $\bm{\pi}_K(\bm{S}_t)=\bm{\pi}_K$. As such, $\pi_i(\bm{S}_{t+1})$ and $m_i^a(\bm{\pi}(\bm{S}_{t+1}))$ are independent of $\bm{A}_t$ and $\bm{S}_t$. It suffices to test the conditional independence of $\widetilde{S}_{i,t+1}$ and $(\bm{A}_t,\bm{S}_t)$ given $(A_{i,t},m_i^a(\bm{A}_t),S_{i,t},m_i^s(\bm{S}_t))$. Here, we adopt the forward-backward learning procedure developed by \cite{shi2020does}. There test does not impose parametric model assumptions on the transition dynamic and is consistent even in high-dimensional settings. In Figure \ref{fig:mf}, we report the distributions of p-values for testing the mean-field approximation assumption associated with two randomly selected regions. It can be seen that the p-values are approximately uniformly distributed. This implies that the mean-field approximation assumption is likely to hold in simulations.}

Our experiments were run on an c5d.9xlarge instance on the AWS EC2 platform, with 36 cores and 72GB RAM.  Computing our estimator for one target policy takes roughly one minute.  
To calculate DR-NS, for each spatial unit, we separately calculate a doubly-robust estimator based on its own state-action-reward triplets. The final estimator is defined as a simple average of all these estimators. 
Mean square errors (MSEs) of value estimators in (a)-(f) are reported in Figure \ref{figure:simu} (b) with different choices of $\sigma_R$ and $T$.  {\color{black}MSEs of DR-NM are larger than $1000$ for all choices of $\sigma_R$ and $T$ and  thus are  not plotted. }

We summarize our findings below: (i) The proposed IS and DR estimators achieve smaller MSEs than DR-NS, DR-NM and Naive Average in all cases. 
In addition, we find that the standard errors of these MSEs are much smaller than the differences between MSEs. Consequently, our methods perform significantly better than the competing baselines. As commented before, this is due to the fact that DR-NS and the naive average estimator ignore interactions among different agents while DR-NM suffers from the curse of dimensionality; (ii) In general, MSEs of our proposed estimators increase with $\sigma_R$ and decreases with $T$, though the trend is not that obvious; (iii) {\color{black}DR outperforms IS and QV in most cases. Although the MSEs of DR and QV are very similar in Figure \ref{figure:simu}, we conduct paired two-sample t-test to test whether the MSE of DR is strictly smaller than that of QV, and find that the tests are significant in most cases. See Appendix \ref{sec:addsimu} of the supplementary article for details. }

\section{Applications}\label{sec:real}
\begin{figure}[!t]
\centering
\subfigure[]{
	\begin{minipage}[t]{0.3\linewidth}
		\centering
		\includegraphics[width=1.0\linewidth]{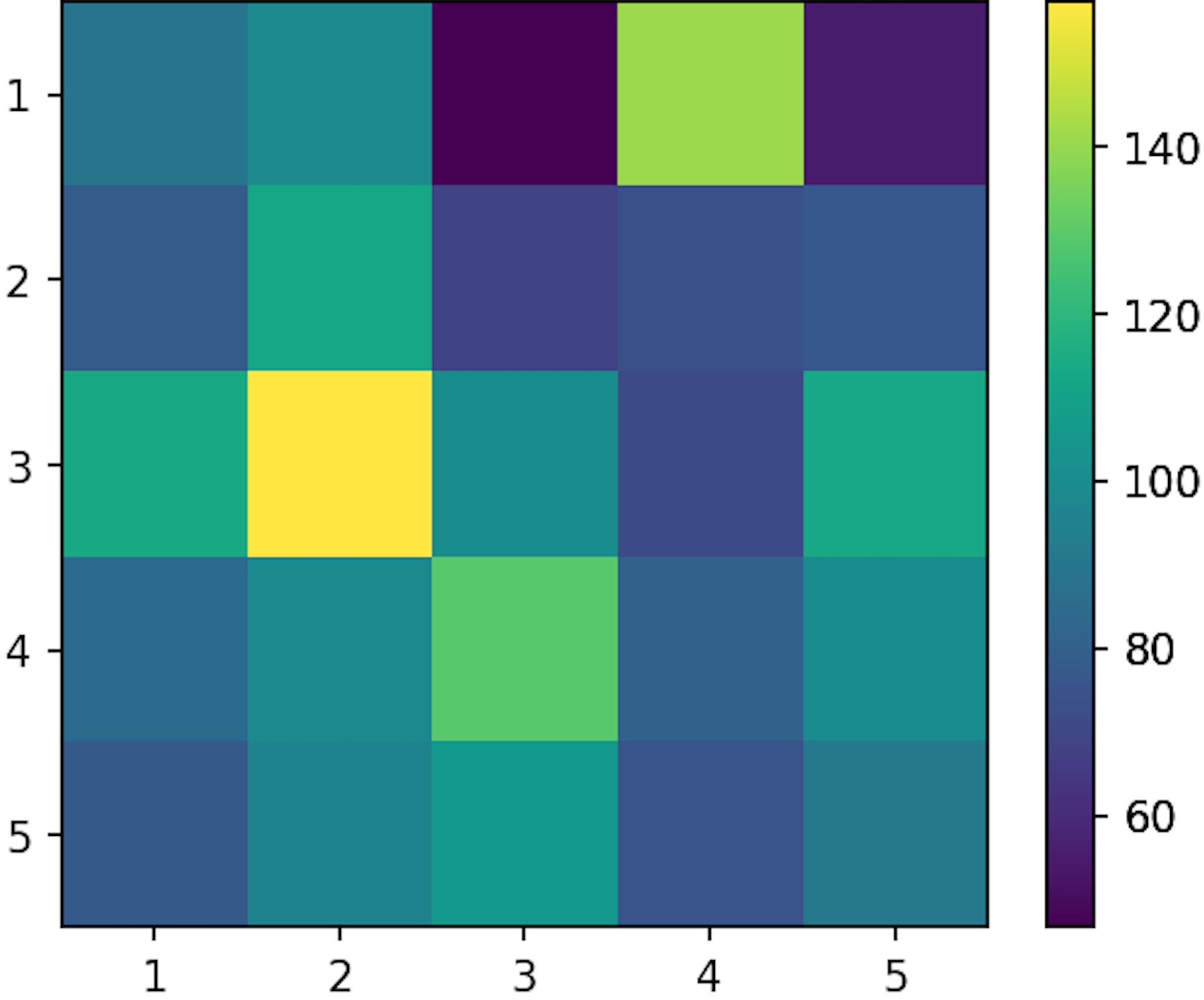}
	\end{minipage}	
}
\subfigure[]{
	\begin{minipage}[t]{0.4\linewidth}
		\centering
		\includegraphics[width=1.0\linewidth]{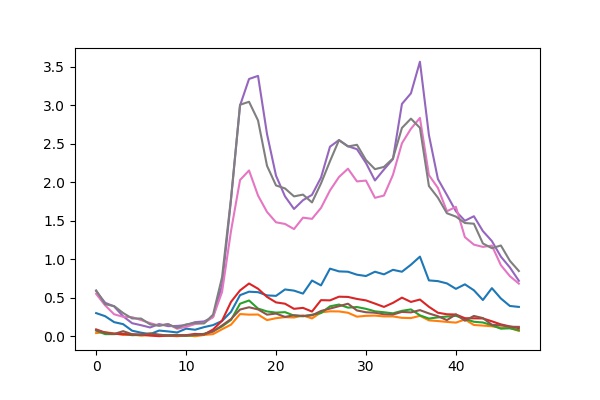}
	\end{minipage}
} %

\caption{(a) visualization of $\mu_i$; (b) 
	number of orders within each spatiotemporal unit, averaged across days. 
	The numbers in (b) 
	are scaled due to sensitivity and privacy concerns. }
\label{fig11}
\end{figure}

We apply the proposed method to a real dataset from a ride-sharing company, to examine the effectiveness
of applying driver-side subsidizing policies and customer recommendation programs. The data is from a randomized experiment, conducted at a given city, partitioned into eight irregular spatial regions, as shown in Figure \ref{fig0}.  Thirty-minutes is defined as one time unit. Figure \ref{fig11} (b) depicts the number of orders $O_{i,t}$ within each spatiotemporal unit, averaged across days. The total Gross Merchandise Volume (GMV) within each spatiotemporal unit is set to the immediate reward $R_{i,t}$. 

Due to data confidentiality, we are not able to use the raw data. Here, we simulate $D_{i,t}$, $O_{i,t}$, $R_{i,t}$ and the degree of mismatch between orders and drivers $M_{i,t}$ for $i=1,\ldots,8$ and  $t=0,\ldots,T-1$. Specifically, we conduct two real data based simulations to evaluate two promotion strategies, one offered to the drivers and another to the passengers. As commented in the introduction, driver-side subsidies will change the spatial distribution of drivers, inducing interference effects in both space and time. Applying  recommendation programs to certain passengers will increase their chances of requesting orders in the future, inducing interference effects in time. 

In the first simulation experiment, we aim to evaluate driver-side subsidizing policy. The numbers of orders $\{O_{i,t}\}_{i,t}$ are independently generated. The spatial distribution of orders are identical across days and the expected number of orders is set to the historical average. The numbers of drivers $\{D_{i,t}\}_{i,t}$ are affected by both the subsidizing action and the spatial distribution of orders. We use a similar generative model as in Section \ref{sec:numerical} with the model parameters estimated using the real dataset. In the second simulation experiment, we aim to evaluate the causal effect of applying passenger recommendation programs. The distribution of $\{O_{i,t}\}_{i,t}$ is set to depend on the action. The distribution of $\{D_{i,t}\}_{i,t}$, however, is set to depend on the orders only. For both experiments, the data generative models are chosen such that the distribution of these generative variables are very similar to that of the observed data. 

{\color{black}Similar to the simulation study, we first apply the forward-backward learning procedure to test the mean-field approximation assumption. Recall that we partition the city into eight spatial regions. The empirical rejection probabilities associated with these regions are 0.03, 0.09, 0.06, 0.07, 0.04, 0.08, 0.03 and 0.1, respectively, under the significance level 0.05. As such, we expect this assumption holds in our data application as well. }

It can be seen from Figure \ref{fig11} that the data are time-varying within each day. To guarantee the transition matrix is homogeneous in time (see MA), we define the state $S_{i,t}=(D_{i,t},O_{i,t},M_{i,t},T_t)^\top$ for any $i$ and $t$, where the variable $T_t$ denotes the time of the day. Since one time unit consists of thirty-minutes, $T_t$ satisfies $T_t=T_{t+48}$ for any $t$. The Markov chain will converge to its limiting distribution under mild conditions as the number of days approaches infinity \citep{lloyd1977reservoirs}. As such, the proposed method remains valid. 
We are interested in evaluating the values under the driver-side policies (i) - (iii) in the first simulation and the passenger-side policies (iv) - (vi) in the second simulation.
\begin{enumerate}
\item[(i)] Driver-side spatially adaptive policy: the four regions that have the largest numbers of orders on average will receive the subsidizing policy all the time. 

\item[(ii)] Driver-side timely adaptive policy: all regions during peak hours will receive the subsidizing policy. 

\item[(iii)] Driver-side spatiotemporally adaptive policy: the four regions that have the largest number of orders on average will receive the subsidizing policy during peak hours. 

\item[(iv)] Passenger-side spatially adaptive policy: the four regions that have the most vacant drivers will receive the subsidizing policy all the time. 

\item[(v)] Passenger-side timely adaptive policy: all regions during time periods with the most vacant drivers will receive the subsidizing policy.

\item[(vi)] Passenger-side spatiotemporally adaptive policy: the four regions that have the most vacant drivers on average will receive the subsidizing policy during non-peak hours. 
\end{enumerate}

\begin{figure}[!t]
\centering
\includegraphics[width=1.0\linewidth]{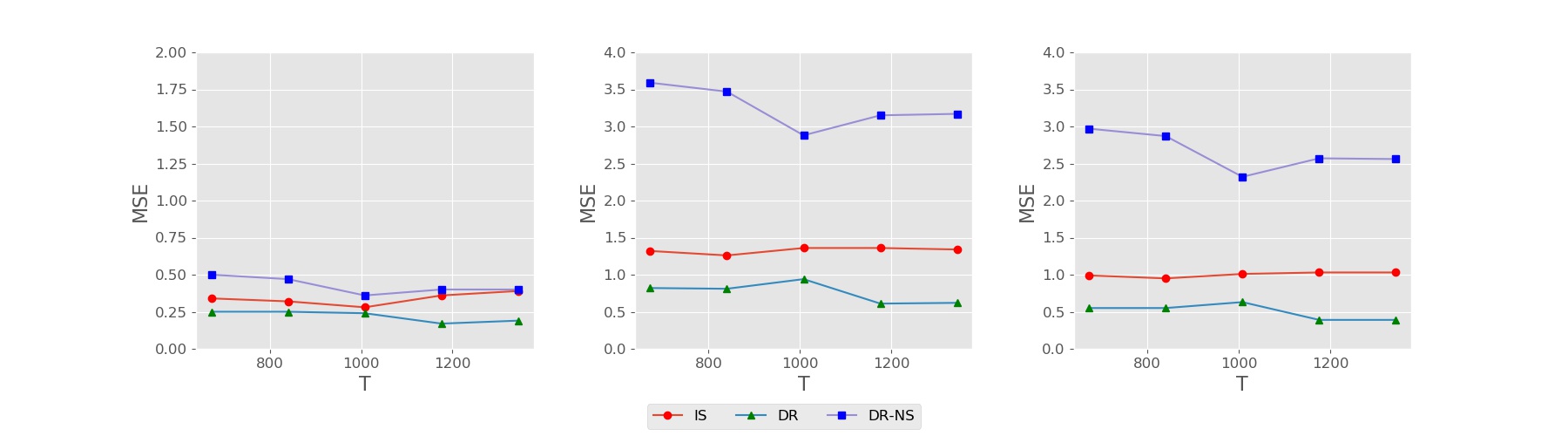}
\caption{Mean squared errors of different value estimates, aggregated over 100 simulations for driver-side subsidizing policies. The target policies are spatially adaptive, timely adaptive and sptiotemporally adaptive, from left to right. {\color{black}MSEs of DR-NM are over 100 in all cases and are hence not reported.}}
\label{fig12}
\end{figure}

\begin{figure}[!t]
\centering
\includegraphics[width=1.0\linewidth]{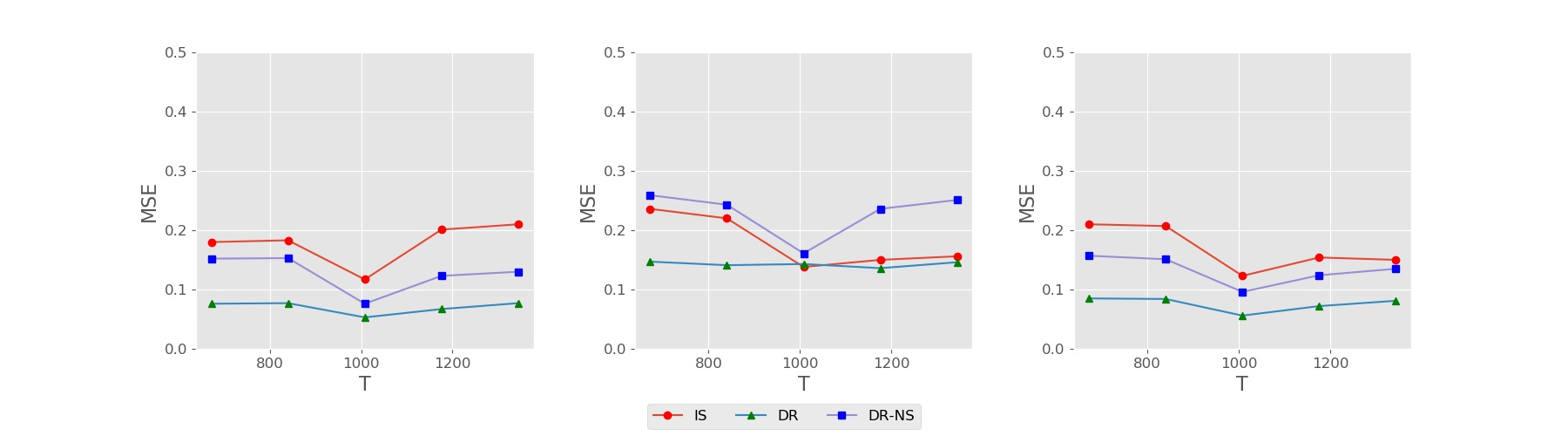}
\caption{Mean squared errors of different value estimates, aggregated over 100 simulations for passenger-side subsidizing policies. The target policies are spatially adaptive, timely adaptive and sptiotemporally adaptive, from left to right.}
\label{fig13}
\end{figure}

For each of the target policy, we compute the MSEs of value estimators constructed based on (a)-(d), with different choices of $T$. The naive average (e) is not considered since the estimator is meaningless when the target differs from the behavior policy. MSEs of DR-NM are over 100 in all cases and are hence not reported. MSEs of other methods are plotted in Figure \ref{fig12} for the first simulation and Figure \ref{fig13} for the second  simulation. It can be seen that DR outperforms other competitors in most cases. It is worth mentioning that DR-NS performs better than IS when evaluating (v) and (vi). This is partly because applying passenger-side recommendation program to a given region only affect the number of orders within that region. The interference effect in space is moderate in this case. 

\section{Discussion}
In this article, we introduce a MARL framework for spatiotemporal treatment effects evaluation and develop a novel off-policy value estimator in MARL. The validity of our method relies on SRA. It would be practically interesting to extend our proposal to settings where SRA is violated. We leave it for future research. 

For simplicity, we set the mean-field function to be the averaged state or action over the neighbors in our numerical experiments. Such a choice works well in our experiments as the MSE of the proposed estimator is much smaller than those of the estimators without considering the spatial dependence or the mean-field approximation. In addition, we find that the resulting mean-field approximation assumption is likely to hold in our numerical experiments. However, it remains unclear whether such a choice of the mean-field function is optimal. We next propose two methods for adaptively selecting the mean-field function in practice. The first method requires a canddiate set of mean-field functions. Given each mean-field function in the candidate set, we can apply existing state-of-the-art conditional independence test to compute its p-value. We next select the mean-field function with the largest p-value. The second method parametrize the mean-field function using some universal approximator (e.g., neural networks) and estimate the associated parameters by minimizing some empirical measures of the conditional dependence of $(\pi_i(\bm{S}_{t+1}), m_i^a(\bm{\pi}(\bm{S}_{t+1})), S_{i,t+1},m_i^s(\bm{S}_{t+1}))$ and $(\bm{A}_t,\bm{S}_t)$ given $(A_{i,t},m_i^a(\bm{A}_t),S_{i,t},m_i^s(\bm{S}_t))$, such as the maximum mean discrepancy \citep[see e.g.,][]{fukumizu2007kernel}. Since the estimated mean-field function minimizes these measures, it is likely to satisfy the conditional independence assumption. It is practically interesting to further investigate these methods. However, it is beyond the scope of the current paper. We leave it for future research. 

\begin{figure}[t]
\centering
\includegraphics[width=6cm]{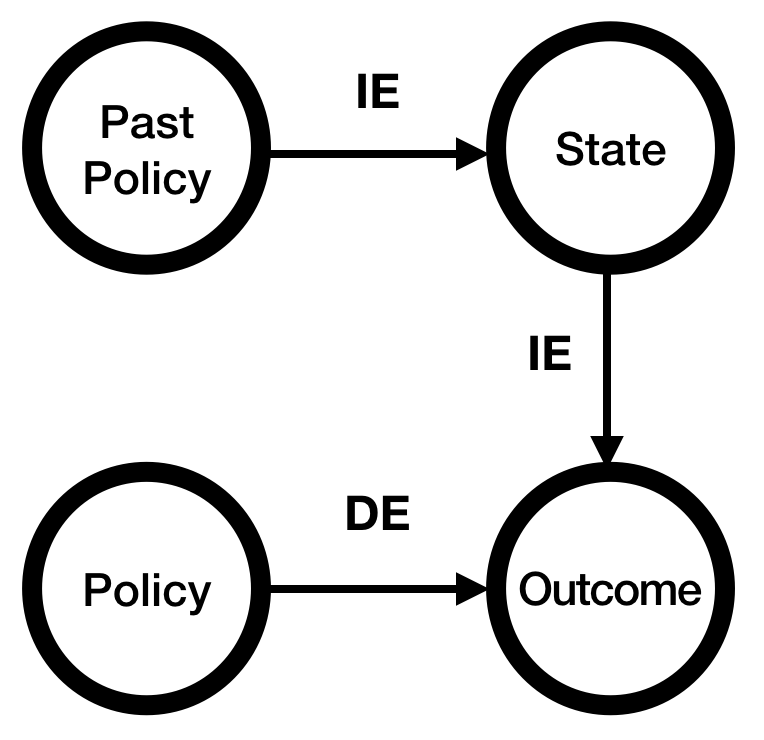}
\caption{Graphical illustration of direct and indirect effects.}\label{DEIE}
\end{figure}

Let $\bm{0}$ and $\bm{1}$ denote two non-dynamic policies that assign Treatments 0 and 1 to each region at any time. We can decompose the value difference $V(\bm{1})-V(\bm{0})$ as the sum of direct effects (DE) and indirect effects (IE). Specifically, it follows from CMIA that
\begin{eqnarray*}
V(\bm{1})-V(\bm{0})=\lim_{t\to \infty} \frac{1}{Nt}\sum_{i=1}^N \sum_{j=0}^t \Mean \{R_{i,j}^*(\bm{1})-R_{i,j}^*(\bm{0})\}\\
=\underbrace{\lim_{t\to \infty} \frac{1}{Nt}\sum_{i=1}^N \sum_{j=0}^t \Mean\{r_i(\bm{1}, \bm{S}_j^*(\bm{0}))-r_i(\bm{0}, \bm{S}_j^*(\bm{0})) \}}_{\hbox{DE}}\\
+\underbrace{\lim_{t\to \infty} \frac{1}{Nt}\sum_{i=1}^N \sum_{j=0}^t \Mean\{r_i(\bm{1}, \bm{S}_j^*(\bm{1}))-r_i(\bm{1}, \bm{S}_j^*(\bm{0})) \}}_{\hbox{IE}},
\end{eqnarray*}
where $\bm{S}_t^*(\bm{0})$ and $\bm{S}_t^*(\bm{1})$ denote the potential states under Policies $\bm{0}$ and $\bm{1}$, respectively.
See Figure \ref{DEIE} for a graphical illustration. The DE represents the sum of the short-term treatment effects on the immediate outcome over time assuming that the baseline policy is being employed in the past.  In contrast,  IE characterizes the carryover effects of past policies that work through the state vector. It is practically interesting to extend the current proposal to estimating DE and IE. However, this is beyond the scope of the current paper. We leave it for future research. 

\bibliography{CausalMARL}
\bibliographystyle{apalike}
\appendix
\section{Theoretical Results}\label{sec:theory}
This section is organized as follows. We first introduce a key lemma that establishes an exponential inequality for the suprema of empirical processes under weak dependence. We next discuss the statistical properties of the proposed estimator.
\subsection{A key lemma}
We briefly introduce the setup before presenting the lemma. Let $\{Z_t:t\ge 0\}$ be a stationary $\beta$-mixing process whose $\beta$-mixing coefficients are given by $\{\beta(q):q\ge 0\}$. 
Let $\mathcal{F}$ be a pointwise measurable class of functions that take $Z_t$ as input with a measurable envelope function $F$. For any $f\in \mathcal{F}$, suppose $\Mean f(Z_0)=0$. Let $\sigma^2>0$ be a positive constant such that $\sup_{f\in \mathcal{F}} \Mean f^2(Z_0)\le \sigma^2 \le \Mean F^2(Z_0)$. We focus providing an exponential inequality for the empirical process $\sup_{f\in \mathcal{F}}|\sum_{t=0}^{T-1} f(Z_t)|$. 

Toward that end, we introduce the notion of the VC type class \citep[][Definition 2.1]{cherno2014}. For any probability measure $Q$, let $e_Q$ denote a semi-metric on $\mathcal{F}$ such that $e_Q(f_1,f_2)=\|f_1-f_2\|_{Q,2} = \sqrt{Q |f_1-f_2|^2}$. An $\epsilon$-net of the space $(\mathcal{F}, e_Q)$ is a subset $\mathcal{F}_{\epsilon}$ of $\mathcal{F}$, such that for every $f\in \mathcal{F}$, there exists some $f_{\epsilon}\in \mathcal{F}_{\epsilon}$ satisfying $e_Q(f,f_{\epsilon}) < \epsilon$. We say that $\mathcal{F}$ is a VC type class with envelope $F$, if there exist constants $c_0 > 0, c_1 \ge 1$, such that $\sup_Q N\left( \mathcal{F},e_Q, \epsilon\|F\|_{Q,2} \right) \le (c_0 / \epsilon)^{c_1}$, for all $0 < \epsilon \le 1$, where the supremum is taken over all finitely discrete probability measures on the support of $\mathcal{F}$, and $\mathbb{N}\left( \mathcal{F},e_Q, \epsilon\|F\|_{Q,2} \right)$ is the infimum of the cardinality of $\epsilon\|F\|_{Q,2}$-nets of $\mathcal{F}$. We refer to $c_1$ as the VC index of $\mathcal{F}$.

\begin{lemma}\label{lemma:EP}
Suppose the envelop function is uniformly bounded by some constant $M>0$. In addition, suppose $\mathcal{F}$ belongs to the class of VC-type class such that $\sup_Q N(\mathcal{F}, e_Q, \varepsilon \|F\|_{Q,2})\le (A/\varepsilon)^{\nu}$ for some $A\ge e,\nu\ge 1$. Then there exist some constants $c,C>0$ such that
\begin{eqnarray*}
	\prob\left(\sup_{f\in \mathcal{F}}\left|\sum_{t=0}^{T-1} f(Z_t)\right|>c\sqrt{\nu q\sigma^2T \log \left(\frac{AM}{\sigma}\right)}+c\nu M \log \left(\frac{AM}{\sigma}\right)+c q\tau+Mq\right)\\
	\le Cq\exp\left(-\frac{\tau^2q}{CT\sigma^2}\right)+Cq\exp\left(-\frac{\tau}{CM}\right)+\frac{T\beta(q)}{q},
\end{eqnarray*}
for any $\tau>0$, $1\le q<T/2$. 
\end{lemma}
As commented in the introduction, Lemma \ref{lemma:EP} is useful for finite-sample analysis of machine learning estimates under weak dependence. It allows us to derive a sharp bound on the difference between the proposed value estimator and the oracle estimator. 

\subsection{Statistical performance guarantees}
To derive these theoretical results, we need one additional condition to characterize the dependence between observations over time.

\smallskip

\noindent (A1) The process $\{(\bm{S}_t,\bm{A}_t)\}_{t\ge 0}$ is strictly stationary. Its $\beta$-mixing coefficients $\{\beta(q)\}_{q\ge 0}$ \citep[see e.g.,][for definition]{Bradley2005} satisfy $\beta(q)\le \kappa_0 \rho^{q}$ for some constants $\kappa_0>0$ and $0<\rho<1$. 

\smallskip

When (A1) holds and the initial distribution of $\{\bm{S}_t\}_{t\ge 0}$ equals to its stationary distribution, the stationarity condition in (A1) is automatically satisfied. The second part of (A1) holds when $\{(\bm{S}_t,\bm{A}_t)\}_{t\ge 0}$ satisfies geometric ergodicity.  Geometric ergodicity is weaker than the uniform ergodicity condition imposed in the existing reinforcement learning literature \citep{bhandari2018finite,zou2019}. 

In addition, let $V_i^*(\bm{\pi})$ and $\omega_i^*$ be the population limit of $\widehat{V}_i(\bm{\pi})$ and $\widehat{\omega}_i$, respectively. We require $V_i^*(\bm{\pi})=V_i(\bm{\pi})$ when \eqref{eqn:MFA2} holds and $\omega_i^*=\omega_i$ when \eqref{eqn:MFA1} holds. 

\begin{thm}\label{thm:oracle}
Suppose (A1) hold, $ NT \mbox{Var}\{\widehat{V}^{\hbox{\scriptsize{DR}}*}(\bm{\pi})\}\to \sigma^2>0$ and $T\to \infty$. Suppose that $\{R_{i,t}, \bar{Q}_i, \omega_{i}^*, V_i(\bm{\pi}):1\le i\le N,t\ge 0\}$ are uniformly bounded from infinity and the set of functions $\{b_i:1\le i\le N\}$ are uniformly bounded from zero. Then as either \eqref{eqn:MFA1} or \eqref{eqn:MFA2} holds, $\sqrt{NT} \{\widehat{V}^{\hbox{\scriptsize{DR}}*}(\bm{\pi})-V(\bm{\pi})\}\stackrel{d}{\to} N(0,\sigma^2)$, where $\stackrel{d}{\to}$ denotes converge in distribution.
\end{thm}

We next investigate the statistical properties of our estimator $\widehat{V}^{\hbox{\scriptsize{DR}}}(\bm{\pi})$. Our theories are generally applicable to a class of density ratio and Q-function estimators that satisfy certain mild conditions. We summarize these conditions in (A2) and  (A3) and present them below. 

\smallskip

\noindent (A2)(i) $\sum_{i=1}^N |V_i^*(\bm{\pi})-\widehat{V}_i(\bm{\pi})|/N=o_p(1)$; (ii) $\widehat{Q}_{i,\bm{\pi}}\in \mathcal{Q}$, $\widehat{\omega}_i\in \mathcal{W}$ almost surely for any $i$ where the function classes $\mathcal{Q}$ and $\mathcal{W}$ belong to VC type classes with VC indices bounded by $\nu=O(T^\kappa)$ for some $0\le \kappa<1$, and envelope functions are bounded by some constant $M$. (iii) $\max_{i,a} \int_{\tilde{s}_i}|\widehat{Q}_{i}(a,\tilde{s}_{i})-Q_{i}^*(a,\tilde{s}_{i})|^2p_{i,b}(\tilde{s}_i)d\tilde{s}_i=o_p(1)$, $\max_i \int_{\tilde{s}_i}|\widehat{\omega}_i(\tilde{s}_{i})-\omega_{i}^*(\tilde{s}_{i})|^2p_{i,b}(\tilde{s}_i)d\tilde{s}_i=o_p(1)$.

\noindent (A3)(i) $\max_i |V_i^*(\bm{\pi})-\widehat{V}_i(\bm{\pi})|^2=o_p((NT)^{-1/2} )$; (ii) $\max_{i,a} \int_{\tilde{s}_i}|\widehat{Q}_{i}(\tilde{s}_{i})-Q_{i}^*(a,\tilde{s}_{i})|^2p_{i,b}(\tilde{s}_i)d\tilde{s}_i=o_p((NT)^{-1/2})$; (iii) $\max_i \int_{\tilde{s}_i}|\widehat{\omega}_i(\tilde{s}_{i})-\omega_{i}^*(\tilde{s}_{i})|^2p_{i,b}(\tilde{s}_i)d\tilde{s}_i=o_p((NT)^{-1/2})$; (iv) $T\gg N\nu^2 \log^4 (NT)$.

Condition (A2) requires $\widehat{V}_i(\bm{\pi})$, $\widehat{Q}_i$ and $\widehat{\omega}_i$ to be consistent whereas (A3) requires these estimators to converge at a certain rate. When using RKHS or neural networks to estimate the Q-function and the density ratio, the corresponding convergence rates have been established \citep{fan2020theoretical,kallus2019efficiently,liao2020batch}. The conditions on $\mathcal{Q}$ and $\mathcal{W}$ in (A2) are mild as these function classes are user-specified. 

\begin{thm}[doubly-robustness]\label{thm:double}
Suppose the conditions in Theorem \ref{thm:oracle} hold. Suppose (A2) holds. Then as either \eqref{eqn:MFA1} or \eqref{eqn:MFA2} holds, we have $\widehat{V}^{\hbox{\scriptsize{DR}}}(\bm{\pi})-V(\bm{\pi})=o_p(1)$. 
\end{thm}

\begin{thm}[oracle property]\label{thm:oracleest}
Suppose the conditions in Theorem \ref{thm:double} hold. Suppose (A3) holds. Then when both  \eqref{eqn:MFA1} and \eqref{eqn:MFA2} hold, we have  $\sqrt{NT} \{\widehat{V}^{\hbox{\scriptsize{DR}}}(\bm{\pi})-V(\bm{\pi})\}\stackrel{d}{\to} N(0,\sigma^2)$.
\end{thm}

{\color{black}Theorem \ref{thm:oracleest} implies that the asymptotic variance of the doubly-robust estimator is the same as that of the oracle estimator, when both mean-field approximations are valid. The explicit formula for $\sigma^2$ is given in Appendix \ref{app:proofthmoracle} of the supplementary article. We also remark that different from the i.i.d. case considered in classical semiparametric statistics, there is no guarantee that the asymptotic variance of the doubly-robust estimator will be smaller than or equal to that of the importance sampling estimator under our setting where observations are time dependent. }




\section{More on the learning procedure}\label{sec:propensity}
We first present a close-form expression for $\widehat{\bm{\beta}}_i$ and $\widehat{\bm{\alpha}}_i$. We next discuss the case where the behavior policy is unknown.
\subsection{Closed-form for $\widehat{\bm{\beta}}_i$}
Notice that $\bm{K}_g$ is symmetric, by some calculations, we obtain
\begin{eqnarray*}
\widehat{\bm{\beta}}_i=(\bm{K}_g\bm{K}_g^\top+T\mu \bm{K}_g)^{-1} \bm{K}_g (\bm{R}+\bm{Q}_i^{*}-\bm{Q}_i-\eta \bm{1})
=(\bm{K}_g+T\mu\bm{I})^{-1} (\bm{R}+\bm{Q}_i^{*}-\bm{Q}_i-\eta \bm{1}).
\end{eqnarray*}
\subsection{Closed-form for $\widehat{\bm{\alpha}}_i$}
Let $\bm{K}_Q$ denotes a $2T\times 2T$ matrix where its $(j_1+1,j_2+1)$th element is given by $K_Q(\widetilde{Z}_{i,j_1}, \widetilde{Z}_{i,j_2})$, we have
\begin{eqnarray*}
Q_i(Z_{i,t})=\bm{\alpha}_i^\top \bm{K}_Q \bm{e}_t\,\,\,\,\hbox{and}\,\,\,\,Q_i(Z_{i,t}^*)=\widehat{\bm{\alpha}}_i^\top \bm{K}_Q \bm{e}_{t+T}.
\end{eqnarray*}
It follow that
\begin{eqnarray*}
\bm{Q}_i^*-\bm{Q}_i=\underbrace{[-\bm{I}_{T}, \bm{I}_{T}]}_{\bm{C}}\bm{K}_Q \widehat{\bm{\alpha}}_i.
\end{eqnarray*}
Note that $\bm{K}_Q$ is symmetric. Let $\bm{E}=\bm{K}_g^\top \{\bm{K}_g+(T-1)\mu \bm{I}\}^{-1}$, $\widehat{\bm{\alpha}}_i$ corresponds to the solution of the following optimization problem, 
\begin{eqnarray*}
\widehat{\bm{\alpha}}_i=\argmin_{\bm{\alpha}} (\bm{R}+\bm{C} \bm{K}_Q \bm{\alpha}-\eta \bm{1})^\top \bm{E}^\top \bm{E} (\bm{R}+\bm{C} \bm{K}_Q \bm{\alpha}-\eta \bm{1})+T \lambda \bm{\alpha}^\top \bm{K}_Q \bm{\alpha}.
\end{eqnarray*}
Taking derivatives with respect to $\bm{\alpha}$ and $\eta$, we obtain
\begin{eqnarray*}
(\widehat{\bm{\alpha}}_i,\widehat{V}_i(\bm{\pi}))^\top=-([\bm{C} \bm{K}_Q,-\bm{1}]^\top \bm{E}^\top \bm{E}[\bm{C} \bm{K}_Q,-\bm{1}]+[T\lambda \bm{K}_Q,\bm{0};\bm{0}^\top,0])^{-1}[\bm{C} \bm{K}_Q,-\bm{1}] \bm{E}^\top \bm{E} \bm{R}.
\end{eqnarray*}
\subsection{Unknown behavior policy}
Note that $b_i(\bm{\pi}|\widetilde{S}_{i,t})=\Mean \{\mathbb{I}(A_{i,t}=\pi_i(\bm{S}_t),m_i^a(\bm{A}_t)=m_i^a(\bm{\pi}(\bm{S}_t)))|\widetilde{S}_{i,t}\}$. It can thus be learned by applying machine learning algorithms to datasets with responses $\{ \mathbb{I}(A_{i,t}=\pi_i(\bm{S}_t),m_i^a(\bm{A}_t)=m_i^a(\bm{\pi}(\bm{S}_t))):0\le t< T \}$ and predictors $\{\widetilde{S}_{i,t}:0\le t<T\}$. 

\section{More on the mean-field approximation}\label{app:moremf}
We first show that \eqref{eqn:MFA2} holds when \eqref{eqn:MFA1} is satisfied and that $(\pi_i(\bm{S}_{t+1}),m_i^a(\bm{\pi}(\bm{S}_{t+1})),\widetilde{S}_{i,t+1})$ is conditionally independent of $(\bm{A}_t,\bm{S}_t)$ given $(A_{i,t},m_i^a(\bm{A}_t),S_{i,t},m_i^s(\bm{S}_t))$. When \eqref{eqn:MFA1} is satisfied, it follows that
\begin{eqnarray*}
	Q_i^{\bm{\pi}}(\bm{a},\bm{s})=\sum_{t\ge 0}\gamma^t \Mean^{\bm{\pi}}\{\bar{r}_i(A_{i,t},m_i^a(\bm{A}_t),\widetilde{S}_{i,t})|\bm{A}_0=\bm{a},\bm{S}_0=\bm{s}\}.
\end{eqnarray*}
It suffices to show that the conditional mean function $\Mean^{\bm{\pi}}\{\bar{r}_i(A_{i,t},m_i^a(\bm{A}_t),\widetilde{S}_{i,t})|\bm{A}_0=\bm{a},\bm{S}_0=\bm{s}\}$ depends on $(\bm{a},\bm{s})$ only through $(a_i,m_i^a(\bm{a}),s_i,m_i^s(\bm{s}))$, for any $t\ge 0$. When $t=0$, this assertion is automatically satisfied by \eqref{eqn:MFA1}. When $t>0$, we have
\begin{eqnarray*}
	\Mean^{\bm{\pi}}\{\bar{r}_i(A_{i,t},m_i^a(\bm{A}_t),\widetilde{S}_{i,t})|\bm{A}_0=\bm{a},\bm{S}_0=\bm{s}\}.
\end{eqnarray*}
It suffices to show that for any square integrable function $h$, $\Mean^{\pi}\{h(\pi_i(\bm{S}_t),m_i^a(\bm{\pi}(\bm{S}_t)),\widetilde{S}_{i,t})|\bm{A}_0=\bm{a},\bm{S}_0=\bm{s}\}$ depends on $(\bm{a},\bm{s})$ only through $(a_i,m_i^a(\bm{a}),s_i,m_i^s(\bm{s}))$, for any $t\ge 1$. 

We next prove this assertion by induction. When $t=1$, this assertion is automatically satisfied by condition. Suppose the assertion holds with $t\le j$. It suffices to show it is satisfied for $t=j+1$. Under MA and the given conditions, for any square integrable function $h$, we have
\begin{eqnarray*} 
	&&\Mean^{\bm{\pi}}\{h(\pi_i(\bm{S}_{j+1}),m_i^a(\bm{\pi}(\bm{S}_{j+1})),\widetilde{S}_{i,j+1})|\bm{A}_0=\bm{a},\bm{S}_0=\bm{s}\}\\
	&=& \Mean^{\bm{\pi}}[\Mean^{\bm{\pi}}\{h(\pi_i(\bm{S}_{j+1}),m_i^a(\bm{\pi}(\bm{S}_{j+1})),\widetilde{S}_{i,j+1})|\bm{A}_j,\bm{S}_j\}|\bm{A}_0=\bm{a},\bm{S}_0=\bm{s}]\\
	&=&\Mean^{\bm{\pi}}[\Mean^{\bm{\pi}}\{h(\pi_i(\bm{S}_{j+1}),m_i^a(\bm{\pi}(\bm{S}_{j+1})),\widetilde{S}_{i,j+1})|A_{i,j},m_i^a(\bm{A}_j),\widetilde{S}_{i,j}\}|\bm{A}_0=\bm{a},\bm{S}_0=\bm{s}]\\
	&=&\Mean^{\bm{\pi}}[\Mean^{\bm{\pi}}\{h(\pi_i(\bm{S}_{j+1}),m_i^a(\bm{\pi}(\bm{S}_{j+1})),\widetilde{S}_{i,j+1})|\pi_i(\bm{S}_j),m_i^a(\bm{\pi}(\bm{S}_{j})),\widetilde{S}_{i,j}\}|\bm{A}_0=\bm{a},\bm{S}_0=\bm{s}].
\end{eqnarray*}
The last line is a function of $(a_i,m_i^a(\bm{a}),s_i,m_i^s(\bm{s}))$ only, given that $\Mean^{\bm{\pi}}\{h(\pi_i(\bm{S}_j),m_i^a(\bm{\pi}(\bm{S}_j)),\widetilde{S}_{i,j}) |\bm{A}_0=\bm{a},\bm{S}_0=\bm{s}\}$ depends on $(\bm{a},\bm{s})$ only through $(a_i,m_i^a(\bm{a}),s_i,m_i^s(\bm{s}))$. This completes the proof. 

\section{Proofs}
We use $c$ and $C$ to denote some generic constants whose values are allowed to vary from place to place. For any two positive sequences $\{a_t\}_{t\ge 1}$ and $\{b_t\}_{t\ge 1}$, we write $a_t\preceq b_t$ if there exists some constant $C>0$ such that $a_t\le Cb_t$ for any $t$. The notation $a_t\preceq 1$ means $a_t=O(1)$.

Lemma \ref{lemma:weight} can thus be proven in a similar manner as Theorem 1 of \cite{liu2018}.  Lemma \ref{lemma:Q} can be similarly proven as Lemma 1 of \cite{shi2020reinforcement}. Theorem \ref{thm:double} can be proven in a similar manner as Theorem \ref{thm:oracleest}. 
In the following, we focus on proving Theorems \ref{thm:oracle}, \ref{thm:oracleest} and Lemma \ref{lemma:EP}. 
\subsection{Proof of Lemma \ref{lemma:EP}}
We break the proof into three steps. In the first step, we use Berbee's coupling lemma \citep[see Lemma 4.1 in][]{Dedecker2002} to approximate $\sup_{f\in \mathcal{F}}|\sum_{t=0}^{T-1} f(Z_t)|$ by sum of i.i.d. variables. In the second step, we apply the tail inequality in Lemma 1 of \cite{Adam2008} to bound the derivation between the empirical process and its mean. Finally, we apply the maximal inequality in Corollary 5.1 of \cite{cherno2014} to bound the expectation of the empirical process. 

\textbf{Step 1.} Following the discussion below Lemma 4.1 of \cite{Dedecker2002},  we can construct a sequence of random variables $\{Z_{t}^0:t\ge 0\}$ such that
\begin{eqnarray}\label{eqn:step1eq1}
	\sup_{f\in \mathcal{F}}\left|\sum_{t=0}^{T-1} f(Z_t)\right|=\sup_{f\in \mathcal{F}}\left|\sum_{t=0}^{T-1} f(Z_t^0)\right|,
\end{eqnarray}
with probability at least $1-T\beta(q)/q$, and that the sequences $\{U_{2i}^0:i\ge 0\}$ and $\{U_{2i+1}^0:i\ge 0\}$ are i.i.d. where $U_i^0=(Z_{iq}^0,Z_{iq+1}^0,\cdots,Z_{iq+q-1}^0)$. 

Recall that $\mathcal{I}_r=\{q\floor{T/q},q\floor{T/q} +1, \cdots,T-1\}$, we have
\begin{eqnarray*}
	\sup_{f\in \mathcal{F}}\left|\sum_{t=0}^{T-1} f(Z_t^0)\right|\le \sum_{j=0}^{q-1}\sup_{f\in \mathcal{F}} \left|\sum_{t=0}^{\floor{T/q}} f(Z_{tq+j}^0)\right|+\sup_{f\in \mathcal{F}}\left|\sum_{t\in \mathcal{I}_r} f(Z_t^0)\right|.
\end{eqnarray*}
Under the boundedness assumption on $F$, the second term on RHS is bounded from above by $Mq$. Without loss of generality, suppose $\floor{T/q}$ is an even number. The first term on the RHS can be bounded from above by $\sum_{j=0}^{2q-1}\sup_{f\in \mathcal{F}} |\sum_{t=0}^{\floor{T/(2q)}} f(Z_{2tq+j}^0)|$. To summarize, we have shown
\begin{eqnarray*}
	\sup_{f\in \mathcal{F}}\left|\sum_{t=0}^{T-1} f(Z_t^0)\right|\le \sum_{j=0}^{2q-1}\sup_{f\in \mathcal{F}} \left|\sum_{t=0}^{\floor{T/(2q)}} f(Z_{2tq+j}^0)\right|+Mq.
\end{eqnarray*}
This together with \eqref{eqn:step1eq1} yields that
\begin{eqnarray}\label{eqn:step1eq2}\\\nonumber
	\prob\left(\sup_{f\in \mathcal{F}}\left|\sum_{t=0}^{T-1} f(Z_t)\right|>2\tau q+Mq\right)\le \prob\left(\sum_{j=0}^{2q-1}\sup_{f\in \mathcal{F}} \left|\sum_{t=0}^{\floor{T/(2q)}} f(Z_{2tq+j}^0)\right|>2\tau q\right)+\frac{T\beta(q)}{q},
\end{eqnarray}
for any $\tau>0$. By Bonferroni's inequality, we obtain
\begin{eqnarray*}
	\prob\left(\sum_{j=0}^{2q-1}\sup_{f\in \mathcal{F}} \left|\sum_{t=0}^{\floor{T/(2q)}} f(Z_{2tq+j}^0)\right|>2\tau q\right)\le \sum_{j=0}^{2q-1} \prob\left(\sup_{f\in \mathcal{F}} \left|\sum_{t=0}^{\floor{T/(2q)}} f(Z_{2tq+j}^0)\right|>\tau \right),
\end{eqnarray*}
for any $\tau>0$. Since the process is stationary, we obtain
\begin{eqnarray*}
	\prob\left(\sum_{j=0}^{2q-1}\sup_{f\in \mathcal{F}} \left|\sum_{t=0}^{\floor{T/(2q)}} f(Z_{2tq+j}^0)\right|>2\tau q\right)\le 2q \prob\left(\sup_{f\in \mathcal{F}} \left|\sum_{t=0}^{\floor{T/(2q)}} f(Z_{2tq}^0)\right|>\tau \right).
\end{eqnarray*}
Combining this together with \eqref{eqn:step1eq2} yields 
\begin{eqnarray}\label{eqn:step1eq3}
	\prob\left(\sup_{f\in \mathcal{F}}\left|\sum_{t=0}^{T-1} f(Z_t)\right|>2\tau q+Mq\right)\le 2q \prob\left(\sup_{f\in \mathcal{F}} \left|\sum_{t=0}^{\floor{T/(2q)}} f(Z_{2tq}^0)\right|>\tau \right)+\frac{T\beta(q)}{q}.
\end{eqnarray}
By construction, $\{Z_{2tq}^0:t\ge 0\}$ are i.i.d. This completes the proof of the first step. 

\textbf{Step 2.} In the second step, we focus on relating the empirical process $\sup_{f\in \mathcal{F}} |\sum_{t=0}^{\floor{T/(2q)}} f(Z_{2tq}^0)|$ to its expectation. Without loss of generality, assume $T=kq$ for some integer $k>0$. Set the constants $\eta$ and $\delta$ in Lemma 1 of \cite{Adam2008} to 1, we obtain
\begin{eqnarray*}
	\prob\left(\sup_{f\in \mathcal{F}} \left|\sum_{t=0}^{\floor{T/(2q)}} f(Z_{2tq}^0)\right|>2\Mean \sup_{f\in \mathcal{F}} \left|\sum_{t=0}^{\floor{T/(2q)}} f(Z_{2tq}^0)\right| +\tau \right)\\
	\le 4\exp\left(-\frac{\tau^2}{2T\sigma^2/q}\right)+\exp\left(-\frac{\tau}{CM}\right),
\end{eqnarray*}
for some constant $C>0$. Combining this together with \eqref{eqn:step1eq3}, we obtain
\begin{eqnarray}\label{eqn:step2}
	\begin{split}
		\prob\left(\sup_{f\in \mathcal{F}}\left|\sum_{t=0}^{T-1} f(Z_t)\right|>4q\Mean \sup_{f\in \mathcal{F}} \left|\sum_{t=0}^{\floor{T/(2q)}} f(Z_{2tq}^0)\right|+2\tau q+Mq\right)\\
		\le 8q\exp\left(-\frac{\tau^2}{2T\sigma^2/q}\right)+2q\exp\left(-\frac{\tau}{CM}\right)+\frac{T\beta(q)}{q},
	\end{split}
\end{eqnarray}
for any $\tau>0$. This completes the proof of the second step.

\textbf{Step 3.} It remains to bound $\Mean \sup_{f\in \mathcal{F}} |\sum_{t=0}^{\floor{T/(2q)}} f(Z_{2tq}^0)|$. By Corollary 5.1 of \cite{cherno2014}, we obtain
\begin{eqnarray*}
	\Mean \sup_{f\in \mathcal{F}} \left|\sum_{t=0}^{\floor{T/(2q)}} f(Z_{2tq}^0)\right|\preceq \sqrt{\frac{\nu \sigma^2T}{q} \log \left(\frac{AM}{\sigma}\right)}+\nu M \log \left(\frac{AM}{\sigma}\right).
\end{eqnarray*}
Combining this together with \eqref{eqn:step2}, we obtain 
\begin{eqnarray*}
	\prob\left(\sup_{f\in \mathcal{F}}\left|\sum_{t=0}^{T-1} f(Z_t)\right|>c\sqrt{\nu q\sigma^2T \log \left(\frac{AM}{\sigma}\right)}+c\nu M \log \left(\frac{AM}{\sigma}\right)+c q\tau+Mq\right)\\
	\le Cq\exp\left(-\frac{\tau^2q}{CT\sigma^2}\right)+Cq\exp\left(-\frac{\tau}{CM}\right)+\frac{T\beta(q)}{q},
\end{eqnarray*}
for some constants $c,C>0$ and any $\tau>0,1\le q<T/2$. The proof is hence completed. 

\subsection{Proof of Theorem \ref{thm:oracle}}\label{app:proofthmoracle}
We introduce some notations. Let $\bar{Q}_{i,t}=\bar{Q}_i(A_{i,t},m_i^a(\bm{A}_t),S_{i,t},m_i^s(\bm{S}_t))$, $\bar{Q}_{i,t}(\bm{\pi})=\bar{Q}_i(\pi_i(\bm{S}_t),\widetilde{S}_{i,t})$ and $\omega_{i,t}^*=\omega_i^*(\widetilde{S}_{i,t})$. 

To prove Theorem \ref{thm:oracle}, we apply the central limit theorem for mixing triangle arrays developed in \cite{francq2005central}. Define $\widehat{V}_t^{\hbox{\scriptsize{DR}}*}(\bm{\pi})$ as
\begin{eqnarray*}
	\frac{1}{N}\sum_{i=1}^N \left[V_i^*(\bm{\pi})+ \omega_{i,t}^*\frac{\mathbb{I}(A_{i,t}=\pi_i(\bm{S}_t),m_i^a(\bm{A}_t)=m_i^a(\bm{\pi}(\bm{S}_t)))}{b_i(\bm{\pi}|\widetilde{S}_{i,t})}
	\{R_{i,t}+\bar{Q}_{i,t+1}(\bm{\pi})-\bar{Q}_{i,t}-V_i^*(\bm{\pi})\}\right],
\end{eqnarray*}
we have $\widehat{V}^{\hbox{\scriptsize{DR}}*}(\bm{\pi})=T^{-1} \sum_{t=0}^{T-1} \widehat{V}_t^{\hbox{\scriptsize{DR}}*}(\bm{\pi})$. Under the stationarity assumption, $\sigma^2$ is equal to the variance of $\sqrt{N}\widehat{V}_t^{\hbox{\scriptsize{DR}}*}(\bm{\pi})$.  

Suppose we have shown each $\widehat{V}_t^{\hbox{\scriptsize{DR}}*}(\bm{\pi})$ is an unbiased estimator for $V(\bm{\pi})$. For $t\in \{0,1,\cdots,T-1\}$, let $x_t=(NT)^{-1/2} \{\widehat{V}_t^{\hbox{\scriptsize{DR}}*}(\bm{\pi})-V(\bm{\pi})\}$. It suffices to show the conditions in (1)-(5) of \cite{francq2005central} hold for $\{x_t:0\le t<T\}$. We next verify these conditions.

\noindent \textbf{Condition (1).} Note that $\{R_{i,t}, \bar{Q}_i, \omega_{i}^*, V_i(\bm{\pi}):1\le i\le N,t\ge 0\}$ are uniformly bounded from infinity, the set of functions $\{b_i:1\le i\le N\}$ are uniformly bounded from zero. As such, $\{x_t:0\le t<T\}$ are uniformly bounded. Condition (1) thus holds for any $\nu^*>0$. 

\noindent \textbf{Condition (2).} This condition is automatically implied by the assumption that $ NT \Var\{\widehat{V}^{\hbox{\scriptsize{DR}}*}(\bm{\pi})\}\to \sigma^2>0$. 

\noindent \textbf{Condition (3).} This condition holds by setting $\kappa=0$ and $T_n=0$ for any $n$.

\noindent \textbf{Condition (4).} Note that the strong mixing coefficients are upper bounded by the $\beta$-mixing coefficients. Under Condition (A4), we can take the sequence $\alpha(h)$ in Condition (4) by $\kappa_0 \rho^h$. 

\noindent \textbf{Condition (5).} Since $\kappa_0 \rho^h$ decays to zero at an exponential rate as $h$ grows to infinity, Condition (5) is automatically satisfied. 

It remains to show $\Mean \widehat{V}_t^{\hbox{\scriptsize{DR}}*}(\bm{\pi})=V(\bm{\pi})$ for any $t$. Suppose (A3) holds. Under the given conditions, we have $V_i^*(\bm{\pi})=V_i(\bm{\pi})$. By Lemma \ref{lemma:Q}, we have
\begin{eqnarray*}
	\Mean \{R_{i,t}+\bar{Q}_{i,t+1}(\bm{\pi})-\bar{Q}_{i,t}-V_i^*(\bm{\pi})|\bm{A}_t,\bm{S}_t\}=0,
\end{eqnarray*}
and hence,
\begin{eqnarray*}
	\Mean \omega_{i,t}^*\frac{\mathbb{I}(A_{i,t}=\pi_i(\bm{S}_t),m_i^a(\bm{A}_t)=m_i^a(\bm{\pi}(\bm{S}_t)))}{b_i(\bm{\pi}|\widetilde{S}_{i,t})}
	\{R_{i,t}+\bar{Q}_{i,t+1}(\bm{\pi})-\bar{Q}_{i,t}-V_i^*(\bm{\pi})\}=0.
\end{eqnarray*}
Consequently, $\Mean \widehat{V}_t^{\hbox{\scriptsize{DR}}*}(\bm{\pi})=N^{-1} \sum_{i=1}^N V_i(\bm{\pi})=V(\bm{\pi})$.

Suppose (A2) holds. Then we have $\omega_{i,t}^*=\omega_{i,t}$ for any $i,t$ where $\omega_{i,t}$ is a shorthand for $\omega_i(\widetilde{S}_{i,t})$. As a result, for any $i,t$, the expectation of the density ratio $\omega_{i,t}^*\mathbb{I}(A_{i,t}=\pi_i(\bm{S}_t),m_i^a(\bm{A}_t)=m_i^a(\bm{\pi}(\bm{S}_t)))/b_i(\bm{\pi}|\widetilde{S}_{i,t})$ equals one. As such, we have
\begin{eqnarray}\label{eqn:proofthm1}
	\begin{split}
		\Mean \left\{V_i^*(\bm{\pi})- \omega_{i,t}^*\frac{\mathbb{I}(A_{i,t}=\pi_i(\bm{S}_t),m_i^a(\bm{A}_t)=m_i^a(\bm{\pi}(\bm{S}_t)))}{b_i(\bm{\pi}|\widetilde{S}_{i,t})}V_i^*(\bm{\pi})\right\}\\
		=V_i^*(\bm{\pi}) \Mean \left\{1- \omega_{i,t}^*\frac{\mathbb{I}(A_{i,t}=\pi_i(\bm{S}_t),m_i^a(\bm{A}_t)=m_i^a(\bm{\pi}(\bm{S}_t)))}{b_i(\bm{\pi}|\widetilde{S}_{i,t})} \right\}=0.
	\end{split}	
\end{eqnarray}
In addition, using similar arguments in \eqref{eqn:valueliu2018}, we have by (A2) that 
\begin{eqnarray}\label{eqn:proofthm11}
	\Mean \left\{\omega_{i,t}^*\frac{\mathbb{I}(A_{i,t}=\pi_i(\bm{S}_t),m_i^a(\bm{A}_t)=m_i^a(\bm{\pi}(\bm{S}_t)))}{b_i(\bm{\pi}|\widetilde{S}_{i,t})}
	R_{i,t}\right\}=V_i(\bm{\pi}).
\end{eqnarray}
Moreover, by some calculations, we have
\begin{eqnarray*}
	&&\Mean \left\{\omega_{i,t}^*\frac{\mathbb{I}(A_{i,t}=\pi_i(\bm{S}_t),m_i^a(\bm{A}_t)=m_i^a(\bm{\pi}(\bm{S}_t)))}{b_i(\bm{\pi}|\widetilde{S}_{i,t})}\bar{Q}_{i,t}\right\}\\&=&\Mean \left\{  \omega_{i,t}^*\frac{\mathbb{I}(A_{i,t}=\pi_i(\bm{S}_t),m_i^a(\bm{A}_t)=m_i^a(\bm{\pi}(\bm{S}_t)))}{b_i(\bm{\pi}|\widetilde{S}_{i,t})}\bar{Q}_{i,t+1}(\bm{\pi})\right\}
	=\int_{\tilde{s}_i} Q_{i}^*(\pi_i(\bm{S}_{t+1}),\tilde{s}_i)p(\bm{\pi},\tilde{s}_i)d\tilde{s}_i.
\end{eqnarray*}
Consequently,
\begin{eqnarray*}
	\Mean \left[\omega_{i,t}^*\frac{\mathbb{I}(A_{i,t}=\pi_i(\bm{S}_t),m_i^a(\bm{A}_t)=m_i^a(\bm{\pi}(\bm{S}_t)))}{b_i(\bm{\pi}|\widetilde{S}_{i,t})}\{\bar{Q}_{i,t+1}(\bm{\pi})-\bar{Q}_{i,t}\}\right]=0.
\end{eqnarray*}
This together with \eqref{eqn:proofthm1} and \eqref{eqn:proofthm11} yields 
\begin{eqnarray*}
	\Mean \left[V_i^*(\bm{\pi})+ \omega_{i,t}^*\frac{\mathbb{I}(A_{i,t}=\pi_i(\bm{S}_t),m_i^a(\bm{A}_t)=m_i^a(\bm{\pi}(\bm{S}_t)))}{b_i(\bm{\pi}|\widetilde{S}_{i,t})}
	\{R_{i,t}+\bar{Q}_{i,t+1}(\bm{\pi})-\bar{Q}_{i,t}-V_i^*(\bm{\pi})\}\right]=V_i(\bm{\pi}). 
\end{eqnarray*}
It follows that $\Mean \widehat{V}^{\hbox{\scriptsize{DR}}*}(\bm{\pi})=V(\bm{\pi})$. 

Thus, $\widehat{V}^{\hbox{\scriptsize{DR}}*}(\bm{\pi})$ is unbiased when either (A2) or (A3) holds. The proof is hence completed. 

\subsection{Proof of Theorem \ref{thm:oracleest}}\label{app:proofthmoracleest}
Similarly, we define $\widehat{Q}_{i,t}$, $\widehat{Q}_{i,t}(\bm{\pi})$ and $\widehat{\omega}_{i,t}$ to be versions of $\bar{Q}_{i,t}$, $\bar{Q}_{i,t}(\bm{\pi})$ and $\omega^*_{i,t}$ with $\bar{Q}_i$ and $\omega_i^*$ replaced with $\widehat{Q}_i$ and $\widehat{\omega}_i$, respectively.  

By Theorem \ref{thm:oracle}, it suffices to show $\sqrt{NT}\widehat{V}^{\hbox{\scriptsize{DR}}}(\bm{\pi})$ is asymptotically equivalent to $\sqrt{NT}\widehat{V}^{\hbox{\scriptsize{DR}}*}(\bm{\pi})$. Note that $\widehat{V}^{\hbox{\scriptsize{DR}}}(\bm{\pi})-\widehat{V}^{\hbox{\scriptsize{DR}}*}(\bm{\pi})$ can be decomposed by $\eta_1+\eta_2+\eta_3+\eta_4+\eta_5$ where
\begin{eqnarray*}
	\eta_1=\frac{1}{NT}\sum_{t=0}^{T-1}\sum_{i=1}^N \left\{\omega_{i,t}^*\frac{\mathbb{I}(A_{i,t}=\pi_i(\bm{S}_t),m_i^a(\bm{A}_t)=m_i^a(\bm{\pi}(\bm{S}_t)))}{b_i(\bm{\pi}|\widetilde{S}_{i,t})}-1\right\}\{V_i^*(\bm{\pi})- \widehat{V}_i(\bm{\pi}) \},\\
	\eta_2=
	\frac{1}{NT}\sum_{t=0}^{T-1}\sum_{i=1}^N \omega_{i,t}^*\frac{\mathbb{I}(A_{i,t}=\pi_i(\bm{S}_t),m_i^a(\bm{A}_t)=m_i^a(\bm{\pi}(\bm{S}_t)))}{b_i(\bm{\pi}|\widetilde{S}_{i,t})}\{\widehat{Q}_{i,t+1}(\bm{\pi})-\widehat{Q}_{i,t}-\bar{Q}_{i,t+1}(\bm{\pi})+\bar{Q}_{i,t} \},\\
	\eta_3=\frac{1}{NT}\sum_{t=0}^{T-1}\sum_{i=1}^N (\widehat{\omega}_{i,t}-\omega_{i,t}^*)\frac{\mathbb{I}(A_{i,t}=\pi_i(\bm{S}_t),m_i^a(\bm{A}_t)=m_i^a(\bm{\pi}(\bm{S}_t)))}{b_i(\bm{\pi}|\widetilde{S}_{i,t})}
	\{R_{i,t}+\bar{Q}_{i,t+1}(\bm{\pi})-\bar{Q}_{i,t}-V_i^*(\bm{\pi})\},\\
	\eta_4=\frac{1}{NT}\sum_{t=0}^{T-1}\sum_{i=1}^N (\widehat{\omega}_{i,t}-\omega_{i,t}^*)\frac{\mathbb{I}(A_{i,t}=\pi_i(\bm{S}_t),m_i^a(\bm{A}_t)=m_i^a(\bm{\pi}(\bm{S}_t)))}{b_i(\bm{\pi}|\widetilde{S}_{i,t})}\{\widehat{Q}_{i,t+1}(\bm{\pi})-\widehat{Q}_{i,t}-\bar{Q}_{i,t+1}(\bm{\pi})+\bar{Q}_{i,t} \},\\
	\eta_5=\frac{1}{NT}\sum_{t=0}^{T-1}\sum_{i=1}^N (\widehat{\omega}_{i,t}-\omega_{i,t}^*)\frac{\mathbb{I}(A_{i,t}=\pi_i(\bm{S}_t),m_i^a(\bm{A}_t)=m_i^a(\bm{\pi}(\bm{S}_t)))}{b_i(\bm{\pi}|\widetilde{S}_{i,t})}\{V_i^*(\bm{\pi})-\widehat{V}_i(\bm{\pi}) \}.
\end{eqnarray*}
In the following, we show $|\eta_j|=o_p((NT)^{-1/2})$, for $j=1,2,\cdots,5$. 

\textbf{Upper bounds on $|\eta_1|$: }Note that $\eta_1=N^{-1}\sum_{i=1}^N \eta_{1,i}$ where
\begin{eqnarray*}
	\eta_{1,i}=\{V_i^*(\bm{\pi})- \widehat{V}_i(\bm{\pi}) \}\left[\frac{1}{T}\sum_{t=0}^{T-1} \left\{\omega_{i,t}^*\frac{\mathbb{I}(A_{i,t}=\pi_i(\bm{S}_t),m_i^a(\bm{A}_t)=m_i^a(\bm{\pi}(\bm{S}_t)))}{b_i(\bm{\pi}|\widetilde{S}_{i,t})}-1\right\}\right].
\end{eqnarray*}
When (A2) holds, we have $\omega_{i,t}^*=\omega_{i,t}$ for any $i,t$. The expectation of the density ratio equals one. As a result, we have
\begin{eqnarray*}
	\Mean \left\{\omega_{i,t}\frac{\mathbb{I}(A_{i,t}=\pi_i(\bm{S}_t),m_i^a(\bm{A}_t)=m_i^a(\bm{\pi}(\bm{S}_t)))}{b_i(\bm{\pi}|\widetilde{S}_{i,t})}-1\right\}=0,
\end{eqnarray*}
for any $i,t$. In the following, we apply the Bernstein's inequality for exponential $\beta$-mixing processes \citep{Chen2015} to bound $|\eta_1|$. 

Under Condition (A4), the $\beta$-mixing coefficients of the sequence
\begin{eqnarray}\label{eqn:set}
	\left\{\omega_{i,t}\frac{\mathbb{I}(A_{i,t}=\pi_i(\bm{S}_t),m_i^a(\bm{A}_t)=m_i^a(\bm{\pi}(\bm{S}_t)))}{b_i(\bm{\pi}|\widetilde{S}_{i,t})}-1:t\ge 0\right\},
\end{eqnarray}
decays to zero at an exponential rate. In addition, all the terms in \eqref{eqn:set} are uniformly bounded by some constant $c>0$. As a result, 
\begin{eqnarray*}
	\max_{t_1,t_2}\Mean \left|\omega_{i,t_1}\frac{\mathbb{I}(A_{i,t_1}=\pi_i(\bm{S}_t),\widetilde{A}_{i,t_1}=m_i^a(\bm{\pi}(\bm{S}_t)))}{b_i(\bm{\pi}|\widetilde{S}_{i,t_1})}-1\right|\left|\omega_{i,t_2}\frac{\mathbb{I}(A_{i,t_2}=\pi_i(\bm{S}_t),\widetilde{A}_{i,t_2}=m_i^a(\bm{\pi}(\bm{S}_t)))}{b_i(\bm{\pi}|\widetilde{S}_{i,t_2})}-1\right|\\=O(1).
\end{eqnarray*}
It thus follows from Theorem 4.2 of \cite{Chen2015} that there exists some constant $C>0$ such that
there exists some constant $C>0$ such that for any $\tau\ge 0$ and integer $1<q<T$,
\begin{eqnarray}\label{prooflemma3eq8}
	\begin{split}
		&\max_i \prob\left(\left|\sum_{t=0}^{T-1} \left\{\omega_{i,t}\frac{\mathbb{I}(A_{i,t}=\pi_i(\bm{S}_t),m_i^a(\bm{A}_t)=m_i^a(\bm{\pi}(\bm{S}_t)))}{b_i(\bm{\pi}|\widetilde{S}_{i,t})}-1\right\} \right|\ge 6\tau \right)\le \frac{T}{q}\beta(q)\\
		+&\max_i \prob\left( \left|\sum_{t\in \mathcal{I}_r} \left\{\omega_{i,t}\frac{\mathbb{I}(A_{i,t}=\pi_i(\bm{S}_t),m_i^a(\bm{A}_t)=m_i^a(\bm{\pi}(\bm{S}_t)))}{b_i(\bm{\pi}|\widetilde{S}_{i,t})}-1\right\}\right|\ge \tau \right)
		+4 \exp\left\{-\frac{\tau^2}{Cq(T+\tau)} \right\},
	\end{split}
\end{eqnarray}
where $\mathcal{I}_r=\{q\floor{T/q},q\floor{T/q} +1, \cdots,T-1\}$. Suppose $\tau\ge qc$. Notice that $|\mathcal{I}_r|\le q$. It follows that
\begin{eqnarray}\label{prooflemma3eq8.5}
	\max_i \prob\left( \left|\sum_{t\in \mathcal{I}_r} \left\{\omega_{i,t}\frac{\mathbb{I}(A_{i,t}=\pi_i(\bm{S}_t),m_i^a(\bm{A}_t)=m_i^a(\bm{\pi}(\bm{S}_t)))}{b_i(\bm{\pi}|\widetilde{S}_{i,t})}-1\right\}\right|\ge \tau \right)=0.
\end{eqnarray}
Under (A4), $\beta(q)=O(\rho^q)$. Set $q=-3\log (NT)/\log \rho$, we obtain $T\beta(q)/q=O(N^{-3} T^{-2})$. Set $\tau=\max\{2\sqrt{CqT\log (NT)}, 4Cq\log(NT)\}$, we obtain as $T\to \infty$ that
\begin{eqnarray*}
	\frac{\tau^2}{2}\ge 2CqT\log (NT)\,\,\,\,\hbox{and}\,\,\,\,\frac{\tau^2}{2}\ge 2Cq\tau \log (nT)\,\,\,\,\hbox{and}\,\,\,\,\tau \ge qc.
\end{eqnarray*} 
Since $\sqrt{CqT\log (NT)}\gg 2Cq\log(NT)$, it follows from \eqref{prooflemma3eq8} and \eqref{prooflemma3eq8.5} that 
\begin{eqnarray*}
	\max_i \prob\left(\left|\sum_{t=0}^{T-1} \left\{\omega_{i,t}\frac{\mathbb{I}(A_{i,t}=\pi_i(\bm{S}_t),m_i^a(\bm{A}_t)=m_i^a(\bm{\pi}(\bm{S}_t)))}{b_i(\bm{\pi}|\widetilde{S}_{i,t})}-1\right\} \right|\ge 12\sqrt{CqT\log (NT)} \right)\preceq N^{-2}T^{-2}. 
\end{eqnarray*}
By Bonferroni's inequality, we obtain the following event occurs with probability at least $1-O(N^{-1} T^{-2})$,
\begin{eqnarray*}
	\max_i \left|\sum_{t=0}^{T-1} \left\{\omega_{i,t}\frac{\mathbb{I}(A_{i,t}=\pi_i(\bm{S}_t),m_i^a(\bm{A}_t)=m_i^a(\bm{\pi}(\bm{S}_t)))}{b_i(\bm{\pi}|\widetilde{S}_{i,t})}-1\right\} \right|\le 12\sqrt{CqT\log (NT)}.
\end{eqnarray*}
It follows that
\begin{eqnarray}\label{eqn:eta1}
	|\eta_1|\le \frac{1}{N}\sum_{i=1}^N |\eta_{1,i}|\preceq \frac{\log (NT)}{\sqrt{T}} \left(\frac{1}{N}\sum_{i=1}^N |V_i^*(\bm{\pi})-\widehat{V}_i(\bm{\pi})|\right),
\end{eqnarray}
with probability approaching 1. Under (A6) and the condition that $T\gg N\log^4 (NT)$, we obtain $\eta_1=o_p((NT)^{-1/2})$. 

\textbf{Upper bounds on $|\eta_2|$: }When (A2) holds, we have $\omega_{i,t}^*=\omega_{i,t}$ for any $i$ and $t$. As discussed in the proof of Theorem \ref{thm:oracle}, we have $\Mean \eta_{2,i}=0$ for any $i$ where
\begin{eqnarray*}
	\eta_{2,i}=
	\frac{1}{T}\sum_{t=0}^{T-1} \omega_{i,t}^*\frac{\mathbb{I}(A_{i,t}=\pi_i(\bm{S}_t),m_i^a(\bm{A}_t)=m_i^a(\bm{\pi}(\bm{S}_t)))}{b_i(\bm{\pi}|\widetilde{S}_{i,t})}\{\widehat{Q}_{i,t+1}(\bm{\pi})-\widehat{Q}_{i,t}-\bar{Q}_{i,t+1}(\bm{\pi})+\bar{Q}_{i,t} \}.
\end{eqnarray*}
In addition, notice that $\eta_{2,i}$ can be written as
\begin{eqnarray*}
	\eta_{2,i}=
	\frac{1}{T}\sum_{t=0}^{T-1} \omega_{i,t}^*\frac{\mathbb{I}(A_{i,t}=\pi_i(\bm{S}_t),m_i^a(\bm{A}_t)=m_i^a(\bm{\pi}(\bm{S}_t)))}{b_i(\bm{\pi}|\widetilde{S}_{i,t})}\{\widehat{Q}_{i,t+1}(\bm{\pi})-\widehat{Q}_{i,t}(\bm{\pi})-\bar{Q}_{i,t+1}(\bm{\pi})+\bar{Q}_{i,t}(\bm{\pi}) \}.
\end{eqnarray*}
We apply Lemma \ref{lemma:EP} to bound $\max_i |\eta_{2,i}|$. Define the class of functions $\mathcal{Q}_{i,\varepsilon}$ by
\begin{eqnarray*}
	\left\{f \in \mathcal{Q}:\max_{i,a} \int_{\tilde{s}_i}|f(a,\tilde{s}_{i})-Q_{i}^*(a,\tilde{s}_{i})|^2p_{i,b}(\tilde{s}_i)d\tilde{s}_i\le \varepsilon \right\},
\end{eqnarray*}
where $\varepsilon=\epsilon N^{-1/2} T^{-1/2}$ for some sufficiently small $\epsilon>0$. It then follows from (A5)(ii) and (iii) that $\widehat{Q}_{i,\bm{\pi}}\in \mathcal{Q}_{\varepsilon}$ for any $i$ with probability tending to $1$. As such, we have
\begin{eqnarray*}
	\eta_{2,i}\le T^{-1} \sup_{Q_i\in \mathcal{Q}_{i,\varepsilon}} \left|\sum_{t=0}^{T-1} \omega_{i,t}^*\frac{\mathbb{I}(A_{i,t}=\pi_i(\bm{S}_t),m_i^a(\bm{A}_t)=m_i^a(\bm{\pi}(\bm{S}_t)))}{b_i(\bm{\pi}|\widetilde{S}_{i,t})}\{f(\pi_i(\bm{S}_{t+1}),\widetilde{S}_{i,t+1})\right.\\
	\left.-f(\pi_i(\bm{S}_{t}),\widetilde{S}_{i,t})-\bar{Q}_{i,t+1}(\bm{\pi})+\bar{Q}_{i,t}(\bm{\pi}) \}\right|.
\end{eqnarray*}

Consider the process $\{(\widetilde{S}_{i,t},A_{i,t},m_i^a(\bm{A}_t),\widetilde{S}_{i,t+1}):t\ge 0\}$. Under (A4), such a process has $\beta$-mixing coefficients $\{\beta^*(q):q\ge 0 \}$ that satisfies $\beta^*(q)=O(\rho^q)$ as well. For any $f$, define the function $g=g(f)$ such that
\begin{eqnarray*}
	g(\widetilde{S}_{i,t},A_{i,t},m_i^a(\bm{A}_t),\widetilde{S}_{i,t+1},\pi_i(\bm{S}_t),\pi_i(\bm{S}_{t+1}))
	=\omega_{i,t}^*\frac{\mathbb{I}(A_{i,t}=\pi_i(\bm{S}_t),m_i^a(\bm{A}_t)=m_i^a(\bm{\pi}(\bm{S}_t)))}{b_i(\bm{\pi}|\widetilde{S}_{i,t})}\\
	\times\{f(\pi_i(\bm{S}_{t+1}),\widetilde{S}_{i,t+1})-f(\pi_i(\bm{S}_t), \widetilde{S}_{i,t})
	-\bar{Q}_{i,t+1}(\bm{\pi})+\bar{Q}_{i,t}(\bm{\pi}) \},
\end{eqnarray*}
almost surely. Consider the class of functions $\mathcal{G}_{i,\varepsilon}=\{g(f):f\in \mathcal{Q}_{i,\varepsilon} \}$. Since $\mathcal{Q}_{i,\varepsilon}$ belongs to the class of VC-type class, so does $\mathcal{G}_{i,\varepsilon}$. Moreover, the VC-index of $\mathcal{G}_{i,\varepsilon}$ is the same as $\mathcal{Q}_{i,\varepsilon}$. Under the boundedness assumption in Theorem \ref{thm:double}, we have
\begin{eqnarray*}
	\Mean g^2(\widetilde{S}_{i,t},A_{i,t},m_i^a(\bm{A}_t),\widetilde{S}_{i,t+1},\pi_i(\bm{S}_t),\pi_i(\bm{S}_{t+1}))\le O(1) \varepsilon,
\end{eqnarray*}
for some constant $O(1)$. In addition, the envelope function of $\mathcal{G}_{i,\varepsilon}$ is uniformly bounded. 

Let $Z_{i,t}=(\widetilde{S}_{i,t},A_{i,t},m_i^a(\bm{A}_t),\widetilde{S}_{i,t+1},\pi_i(\bm{S}_t),\pi_i(\bm{S}_{t+1}))$. Applying Lemma \ref{lemma:EP}, we obtain 
\begin{eqnarray*}
	\max_i \prob\left(\sup_{g\in \mathcal{G}_{i,\varepsilon}}\left|\sum_{t=0}^{T-1} g(Z_{i,t})\right|>c\sqrt{\nu q\varepsilon T \log \left(\frac{1}{\varepsilon}\right)}+c\nu \log \left(\frac{1}{\varepsilon}\right)+c q\tau+c q\right)\\
	\le c q\exp\left(-\frac{\tau^2q}{cT\varepsilon }\right)+cq\exp\left(-\frac{\tau}{c}\right)+\frac{T\beta(q)}{q},
\end{eqnarray*}
for some constant $c>0$. Set $q=-2\log (NT)/\log \rho$, we have $T\beta(q)/q=O(N^{-2} T^{-1})$. Set $\tau=\max(2c\log (NT), \sqrt{2c\varepsilon T\log (NT)/q})$, the RHS is bounded by $O(N^{-2} T^{-1} \log (NT))$. By Bonferroni's inequality, we obtain with probability tending to $1$ that
\begin{eqnarray*}
	T|\eta_{2,i}|\le c\sqrt{\nu q\varepsilon T \log \left(\frac{1}{\varepsilon}\right)}+c\nu \log \left(\frac{1}{\varepsilon}\right)+c q\tau+c q,\,\,\,\,\forall i\in \{1,\cdots,N\},
\end{eqnarray*}
or equivalently,
\begin{eqnarray*}
	\max_i |\eta_{2,i}|\preceq \sqrt{\frac{\epsilon}{NT}}+o\left(\frac{1}{\sqrt{NT}}\right),
\end{eqnarray*}
under the condition that $T\gg N \nu^2 \log^4 (NT)$. Since $\varepsilon$ can be chosen arbitrarily small, we obtain $\max_i |\eta_{2,i}|=o_p((NT)^{-1/2})$. This in turn implies $\eta_2=o_p((NT)^{-1/2})$.

\textbf{Upper bounds on $|\eta_3|$: }Using similar arguments in proving $\eta_2=o_p((NT)^{-1/2})$, we can show $\eta_3=o_p((NT)^{-1/2})$. We omit the technical details to save space. 

\textbf{Upper bounds on $|\eta_4|$ and $|\eta_5|$: }We show $\eta_4=o_p((NT)^{-1/2})$ only. Using similar arguments, one can show $\eta_5=o_p((NT)^{-1/2})$. 

Note that
\begin{eqnarray*}
	\eta_4=\frac{1}{NT}\sum_{t=0}^{T-1}\sum_{i=1}^N (\widehat{\omega}_{i,t}-\omega_{i,t}^*)\frac{\mathbb{I}(A_{i,t}=\pi_i(\bm{S}_t),m_i^a(\bm{A}_t)=m_i^a(\bm{\pi}(\bm{S}_t)))}{b_i(\bm{\pi}|\widetilde{S}_{i,t})}\\
	\times \{\widehat{Q}_{i,t+1}(\bm{\pi})-\widehat{Q}_{i,t}(\bm{\pi})-\bar{Q}_{i,t+1}(\bm{\pi})+\bar{Q}_{i,t}(\bm{\pi}) \}\\
	\le O(1) \frac{1}{NT}\sum_{t=0}^{T-1}\sum_{i=1}^N |\widehat{\omega}_{i,t}-\omega_{i,t}^*||\widehat{Q}_{i,t+1}(\bm{\pi})-\widehat{Q}_{i,t}(\bm{\pi})-\bar{Q}_{i,t+1}(\bm{\pi})+\bar{Q}_{i,t}(\bm{\pi})|\\
	\le O(1) \left\{ \frac{1}{NT}\sum_{t=0}^{T-1}\sum_{i=1}^N [(\widehat{\omega}_{i,t}-\omega_{i,t}^*)^2+\{\widehat{Q}_{i,t+1}(\bm{\pi})-\widehat{Q}_{i,t}(\bm{\pi})-\bar{Q}_{i,t+1}(\bm{\pi})+\bar{Q}_{i,t}(\bm{\pi})\}^2] \right\}\\
	\le O(1) \left\{ \frac{1}{NT}\sum_{t=0}^{T-1}\sum_{i=1}^N (\widehat{\omega}_{i,t}-\omega_{i,t}^*)^2\right\}+O(1) \left\{ \frac{1}{NT}\sum_{t=0}^{T-1}\sum_{i=1}^N\{\widehat{Q}_{i,t}(\bm{\pi})- \bar{Q}_{i,t}(\bm{\pi})\}^2\right\},
\end{eqnarray*}
where $O(1)$ denotes some universal constant, and the last two inequalities are due to Cauchy-Schwarz inequality. 

To prove $\eta_4=o_p((NT)^{-1/2})$, it suffices to show
\begin{eqnarray}\label{eqn:Qsq}
	\max_i \left[\frac{1}{T} \sum_{t=0}^{T-1}\{\widehat{Q}_{i,t}(\bm{\pi})-\bar{Q}_{i,t}(\bm{\pi})\}^2 \right]=o_p((NT)^{-1/2}),
\end{eqnarray}
and
\begin{eqnarray}\label{eqn:omegasq}
	\max_i \left\{ \frac{1}{T} \sum_{t=0}^{T-1}(\widehat{\omega}_{i,t}-\omega_{i,t}^*)^2 \right\}=o_p((NT)^{-1/2}).
\end{eqnarray}

The left-hand-side (LHS) of \eqref{eqn:Qsq} can be upper bounded by
\begin{eqnarray*}
	\max_i \sup_{f\in \mathcal{Q}_{i,\varepsilon}} \left[\frac{1}{T}\sum_{t=0}^{T-1} \{f(\pi_i(\bm{S}_t),\widetilde{S}_{i,t})-\bar{Q}_{i,t}(\bm{\pi})\}^2 \right],
\end{eqnarray*}
with probability tending to $1$. 
Using similar arguments in proving $\eta_2=o_p((NT)^{-1/2})$, we can show
\begin{eqnarray*}
	\max_i \sup_{f\in \mathcal{Q}_{i,\varepsilon}} \left|\frac{1}{T}\sum_{t=0}^{T-1} \{f(\pi_i(\bm{S}_t),\widetilde{S}_{i,t})-\bar{Q}_{i,t}(\bm{\pi})\}^2-\frac{1}{T}\sum_{t=0}^{T-1} \Mean \{f(\pi_i(\bm{S}_t),\widetilde{S}_{i,t})-\bar{Q}_{i,t}(\bm{\pi})\}^2  \right|\\\preceq \frac{\epsilon}{\sqrt{NT}}+o\left(\frac{1}{\sqrt{NT}}\right),
\end{eqnarray*} 
with probability tending to $1$. Under (A6), we have
\begin{eqnarray*}
	\max_i \sup_{f\in \mathcal{Q}_{i,\varepsilon}} \left|\frac{1}{T}\sum_{t=0}^{T-1} \Mean \{f(\pi_i(\bm{S}_t),\widetilde{S}_{i,t})-\bar{Q}_{i,t}(\bm{\pi})\}^2  \right|\preceq \frac{\epsilon}{\sqrt{NT}}.
\end{eqnarray*} 
It follows that 
\begin{eqnarray*}
	\max_i \sup_{f\in \mathcal{Q}_{i,\varepsilon}} \left[\frac{1}{T}\sum_{t=0}^{T-1} \{f(\pi_i(\bm{S}_t),\widetilde{S}_{i,t})-\bar{Q}_{i,t}(\bm{\pi})\}^2 \right]\preceq \frac{\epsilon}{\sqrt{NT}}+o\left(\frac{1}{\sqrt{NT}}\right),
\end{eqnarray*}
with probability tending to $1$. Let $\epsilon\to 0$, we obtain \eqref{eqn:Qsq}. Similarly, we can show \eqref{eqn:omegasq} holds. The proof is hence completed. 

\section{Additional Simulation Results}\label{sec:addsimu}
\begin{figure}[!t]
	\includegraphics[width=10cm]{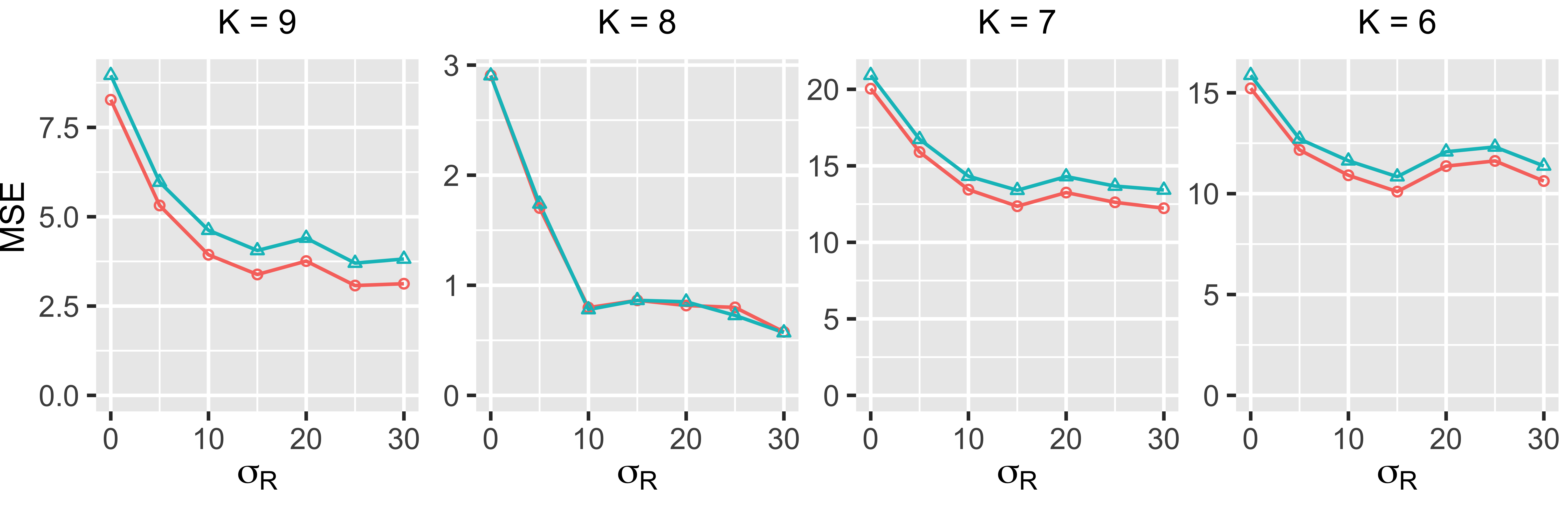}
	\includegraphics[width=10cm]{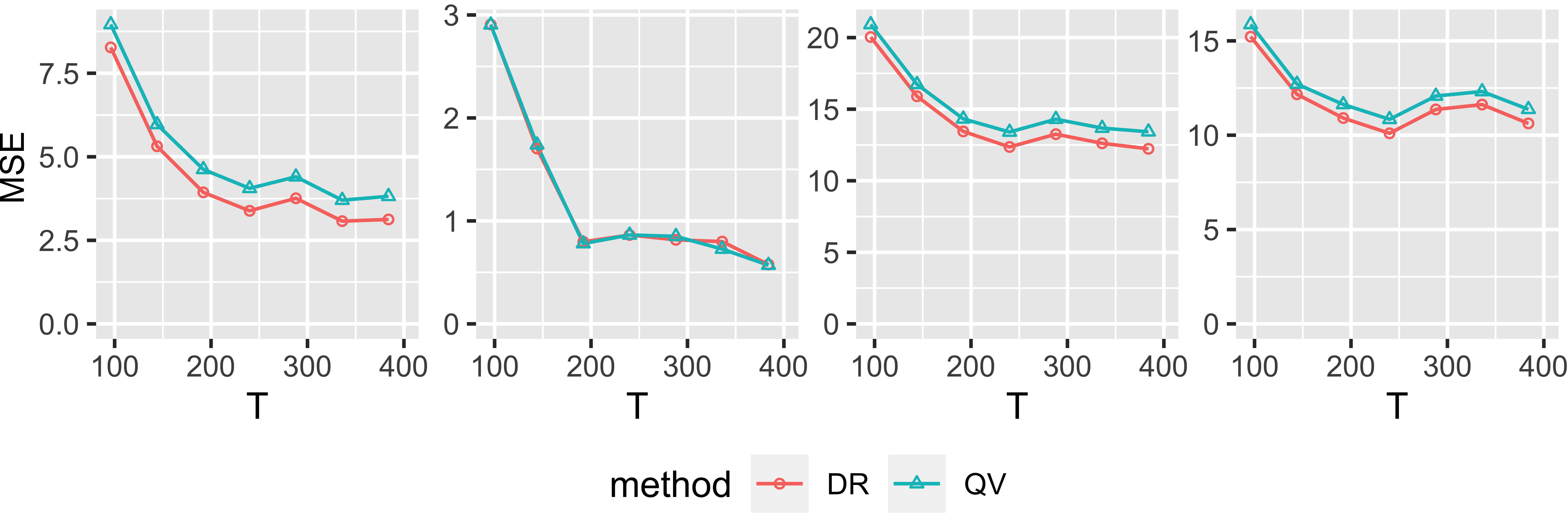}
	\caption{Mean squared errors of the DR and QV estimators, aggregated over 100 simulations. $T$ is set to $336$ (the experiment lasts for two weeks and each hour is treated as one time unit) in the top plots and $\sigma_R$ is set to $15$ in the bottom plots.
	}\label{fig:zoom-in}
\end{figure}

We conduct paired two-sample t-test to test whether the MSE of DR is strictly smaller than that of QV in our simulation studies. The p-values of these t-tests are reported in Tables \ref{tab:pvalue1} and \ref{tab:pvalue2}. It can be seen that most p-values are significant under the 0.05 significance level. Figure \ref{fig:zoom-in} plots the mean squared errors (MSEs) of the DR and QV estimators with a different y-axis scale. It can be clearly seen that our estimator achieves a strictly smaller MSE when $K=6,7$ or $9$. 

\begin{table}[t]
	\begin{tabular}{cccccccc}
		\toprule
		& $\sigma_R=0$ & $\sigma_R=5$ & $\sigma_R=10$ & $\sigma_R=15$ & $\sigma_R=20$ & $\sigma_R=25$ & $\sigma_R=30$ \\ \midrule
		$K=9$ & $1.1\times 10^{-116}$ & $1.3\times 10^{-102}$ & $5.3\times 10^{-73}$ & $8.4\times 10^{-40}$ & $1.2\times 10^{-22}$ & $2.8\times 10^{-14}$ & $2.3\times 10^{-8}$  \\
		$K=8$ & $0.50$ & $0.71$ & $0.97$ & $0.98$ & $0.99$ & $1.00$ & $1.00$  \\
		$K=7$ & $6.6\times 10^{-87}$ & $1.9\times 10^{-75}$ & $3.4\times 10^{-41}$ & $1.2\times 10^{-27}$ & $4.9\times 10^{-19}$ & $4.1\times 10^{-15}$ & $6.4\times 10^{-7}$ \\
		$K=6$ & $8.8\times 10^{-43}$ & $1.5\times 10^{-39}$ & $1.2\times 10^{-19}$ & $3.6\times 10^{-13}$ & $8.8\times 10^{-10}$ & $9.8\times 10^{-4}$ & $5.8\times 10^{-3}$ \\
		\bottomrule
	\end{tabular}	
	\caption{P-values of paired two-sample t-test in the simulation study with different combinations of $\sigma_R$ and $K$. $T$ is fixed to 336.}\label{tab:pvalue1}
\end{table}	
\begin{table}	
	\begin{tabular}{cccccccc}
		\toprule
		& $\sigma_R=0$ & $\sigma_R=5$ & $\sigma_R=10$ & $\sigma_R=15$ & $\sigma_R=20$ & $\sigma_R=25$ & $\sigma_R=30$ \\ \midrule
		$K=9$ & $1.1\times 10^{-116}$ & $1.3\times 10^{-102}$ & $5.3\times 10^{-73}$ & $8.4\times 10^{-40}$ & $1.2\times 10^{-22}$ & $2.8\times 10^{-14}$ & $2.3\times 10^{-8}$  \\
		$K=8$ & $0.50$ & $0.71$ & $0.97$ & $0.98$ & $0.99$ & $1.00$ & $1.00$  \\
		$K=7$ & $6.6\times 10^{-87}$ & $1.9\times 10^{-75}$ & $3.4\times 10^{-41}$ & $1.2\times 10^{-27}$ & $4.9\times 10^{-19}$ & $4.1\times 10^{-15}$ & $6.4\times 10^{-7}$ \\
		$K=6$ & $8.8\times 10^{-43}$ & $1.5\times 10^{-39}$ & $1.2\times 10^{-19}$ & $3.6\times 10^{-13}$ & $8.8\times 10^{-10}$ & $9.8\times 10^{-4}$ & $5.8\times 10^{-3}$ \\
		\bottomrule
	\end{tabular}	
	\caption{P-values of paired two-sample t-test in the simulation study with different combinations of $T$ and $K$. $\sigma_R$ is fixed to 15.}\label{tab:pvalue2}
\end{table}

\end{document}